\documentclass{article}

\usepackage[preprint]{neurips_2025}

\usepackage[utf8]{inputenc} % allow utf-8 input
\usepackage[T1]{fontenc}    % use 8-bit T1 fonts
\usepackage{hyperref}       % hyperlinks
\usepackage{url}            % simple URL typesetting
\usepackage{booktabs}       % professional-quality tables
\usepackage{amsfonts}       % blackboard math symbols
\usepackage{nicefrac}       % compact symbols for 1/2, etc.
\usepackage{wrapfig}
\usepackage[most]{tcolorbox}
\usepackage{caption}
\usepackage{lmodern}
\usepackage{multirow}
\usepackage{graphicx}
\usepackage{amsmath}
\usepackage{subcaption}
\usepackage{array}     % For column widths
\usepackage{makecell}  
\usepackage{microtype}      % microtypography
\usepackage{xcolor}         % colors
\usepackage[inline]{enumitem}

\definecolor{brown}{rgb}{0.6, 0.4, 0.2}

\title{\textsc{FABLE}: A Novel Data-\underline{F}low \underline{A}nalysis \underline{B}enchmark on Procedural Text for \underline{L}arge Language Model \underline{E}valuation}

\author{%
  Vishal Pallagani \\
  \small{University of South Carolina} \\
  \texttt{vishalp@mailbox.sc.edu} \\
  \And
    Nitin Gupta \\
  \small{University of South Carolina} \\
  \texttt{niting@email.sc.edu} \\
    \AND
    John Aydin \\
  \small{University of South Carolina} \\
  \texttt{jaaydin@email.sc.edu} \\
  \And
  Biplav Srivastava \\
  \small{University of South Carolina} \\
  \texttt{biplav.s@sc.edu} \\
}

\begin{document}

\maketitle

\begin{abstract}
% \biplav{Better to have name with D (for data) emphasized. Like: DIMPLE - \underline{D}ata-Flow Analys\underline{i}s Bench\underline{m}ark on \underline{P}rocedural Text for \underline{L}arge Language Model \underline{E}valuation;  Changed abstract slightly.} 
Understanding how data moves, transforms, and persists, known as \textit{data flow}, is fundamental to reasoning in procedural tasks. Despite their fluency in natural and programming languages, large language models (LLMs), although increasingly being applied to decisions with procedural tasks, have not been systematically evaluated for their ability to perform data-flow reasoning. We introduce \textsc{FABLE}, an extensible benchmark designed to assess LLMs' understanding of data flow using structured, procedural text. \textsc{FABLE} adapts eight classical data-flow analyses from software engineering: reaching definitions, very busy expressions, available expressions, live variable analysis, interval analysis, type-state analysis, taint analysis, and concurrency analysis. These analyses are instantiated across three real-world domains: cooking recipes, travel routes, and automated plans. The benchmark includes 2,400 question–answer pairs, with 100 examples for each domain-analysis combination. We evaluate three types of LLMs: a reasoning-focused model (\texttt{deepseek-r1:8b}), a general-purpose model (\texttt{llama3.1:8b}), and a code-specific model (\texttt{granite-code:8b}). Each model is tested using majority voting over five sampled completions per prompt. Results show that the reasoning model achieves higher accuracy, but at the cost of over 20x average slowdown in inference time compared to the other models. On the contrary, the general-purpose and code-specific models perform close to random chance. \textsc{FABLE} provides the first diagnostic benchmark to systematically evaluate data-flow reasoning and offers insights for developing models with stronger procedural understanding. \href{https://github.com/VishalPallagani/FABLE/}{
  \includegraphics[height=0.9em]{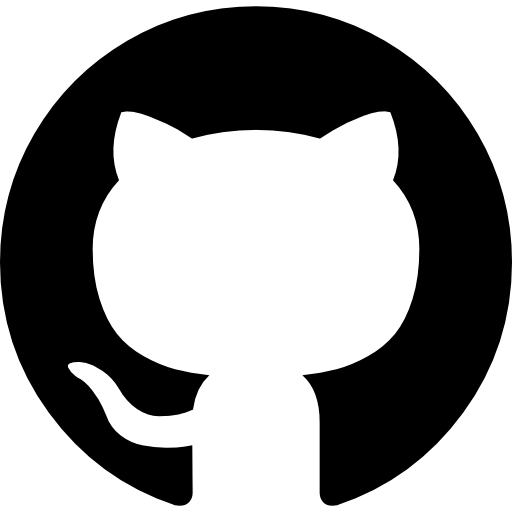}
} 
\href{https://huggingface.co/datasets/g-nitin/FABLE}{
  \includegraphics[height=0.9em]{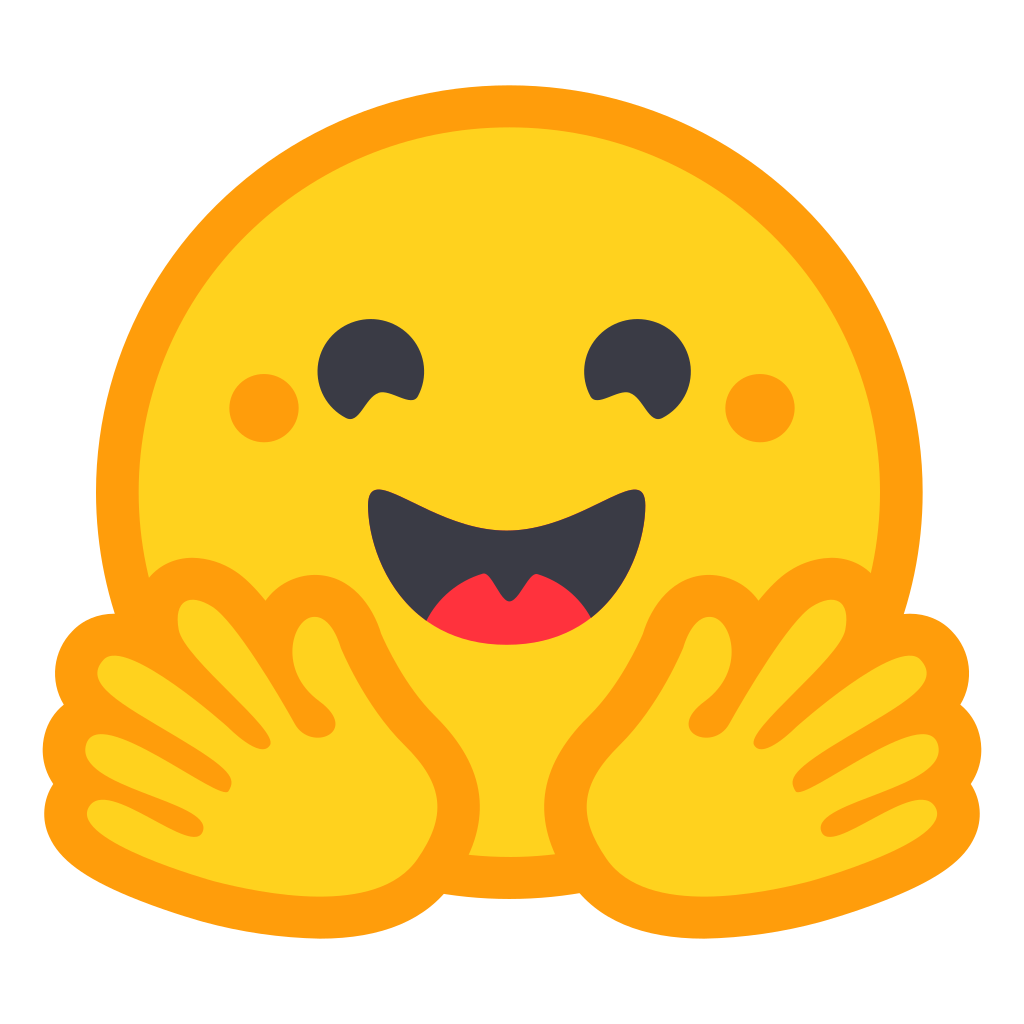}
}

\end{abstract}

\section{Introduction}

Procedural texts, such as recipes, travel routes, and plans, describe dynamic processes that unfold over time. Understanding these texts requires models to reason not only about individual statements but also about how information evolves throughout a sequence of actions. A core capability underlying this understanding is data-flow reasoning, which involves tracking how entities, variables, and their attributes are defined, modified, and propagated as a procedure progresses. 

In software engineering, data-flow analysis \cite{dwyer1994data, fosdick1976data, kennedy1979survey, pollock2002incremental} plays a foundational role in verifying the correctness of programs by examining how variables change across control structures. This capability is equally essential for comprehending procedural language, where similar patterns of definition and modification occur in naturalistic contexts. For example, a model reading a recipe must track ingredient transformations and temporal dependencies to maintain a coherent understanding of the process. Despite this alignment between code and procedural text, current evaluation benchmarks do not fully test such reasoning capabilities in LLMs.

In literature, researchers have introduced datasets that evaluate isolated aspects of procedural reasoning. Some focus on tracking entity states \cite{tandon2018reasoning}, while others target causal or temporal relationships between steps \cite{tandon2019wiqa, ning-etal-2020-torque}. These benchmarks have highlighted the importance of procedural understanding, but they also expose significant gaps. For instance, on the WIQA dataset \cite{tandon2019wiqa}, transformer-based models reached only about 74\% accuracy compared to 96\% for humans. Similarly, on the temporal ordering benchmark TORQUE \cite{ning-etal-2020-torque}, models lag by approximately 30\% behind human performance. Moreover, existing benchmarks often isolate a single reasoning dimension, such as causality or time, and tend to use short, domain-specific descriptions. In contrast, real-world procedures involve multiple interdependent factors, including causality, temporal structure, and entity state evolution over extended sequences.

In this paper, we introduce \textsc{FABLE}, a benchmark specifically designed to evaluate data-flow reasoning in LLMs through procedural texts. \textsc{FABLE} is inspired by program analysis techniques and extends them to natural language. It emphasizes comprehensive tracking of ``where and how information flows'' through a procedure, bridging the gap between causal effects, entity state changes, and temporal dependencies within the same benchmark. \textsc{FABLE} adapts eight classical data-flow analyses from software engineering, including reaching definitions, available expressions, very busy expressions, live variable analysis, interval analysis, type-state analysis, taint analysis, and concurrency analysis. These analyses are instantiated across three real-world domains: cooking recipes, travel routes, and automated plans. Each analysis challenges models to identify how information is introduced, propagated, and transformed over time. The benchmark contains 2,400 question-and-answer pairs, with 100 examples for each analysis within each domain.

We evaluate three categories of large language models: a reasoning-focused model, a general-purpose model, and a code-specific model. Each model is tested using majority voting over five sampled completions per prompt. Our results show that the reasoning model achieves the highest accuracy but has over 20x average slowdown in inference time compared to the other two models. In contrast to the reasoning model, the general-purpose and code-specific models perform close to random chance. These findings reinforce the challenges of data-flow reasoning and demonstrate the utility of \textsc{FABLE} as a diagnostic benchmark. \textbf{Our key contributions are as follows:}
\begin{enumerate*}[label=\textbf{(\arabic*)}]
    \item We introduce \textsc{FABLE}, the first benchmark to adapt eight classical data-flow analyses from software engineering to procedural text, enabling comprehensive evaluation of data-flow reasoning in LLMs; 
    \item We benchmark three categories of large language models and conduct a two-fold empirical analysis: (i) we evaluate to show that the dataset presents a substantial reasoning challenge, and (ii) we establish baseline performance and computational tradeoffs across reasoning-focused, general-purpose, and code-specific LLMs.
\end{enumerate*}

\section{Background \& Related Works}

\subsection{Procedural Text and the Need for Data-Flow Understanding}
We refine the definition of procedural text that is defined in prior literature \cite{zhang-etal-2023-causal}. A \emph{procedural text} is a triple $\mathcal{P} \;=\; (\,G,\; S,\; \preceq\,),$ where: $G$ is a \emph{goal}, expressed in natural language describing the final desired outcome, $S = \{\, s_1,\dots, s_n \}$ is a finite set of \emph{steps}, each step $s_i$ being a natural-language instruction describing some action or sub-procedure to perform, and $\preceq$ is a (partial or total) order on $S$, indicating the permissible sequence(s) in which these steps can be executed to achieve $G$. Procedural texts are ubiquitous in the real world. They appear in domains as diverse as cooking, scientific experimentation, travel planning, and robotics. Their role is central to knowledge sharing, decision-making, automation, and task execution in both human and machine systems. Understanding procedural texts requires tracking how entities and their attributes evolve across ordered steps, how one step's output serves as another's input, and how the process collectively satisfies the intended goal \cite{tandon-etal-2018-reasoning}. This naturally calls for data-flow reasoning, a concept rooted in program analysis, where variable definitions and their propagation through execution paths are systematically tracked. Analogously, in natural language, such reasoning involves identifying where entities are introduced, how they are transformed, and which steps depend on prior states. Despite its importance, existing benchmarks only target isolated aspects of procedural reasoning, as discussed below.

\subsection{Prior Relevant Benchmarks}

Early work on procedural text understanding focused on tracking how entities change state throughout a process. A comparison of \textsc{FABLE} with relevant prior benchmarks is presented in Table~\ref{tab:procedural-benchmarks}. ProPara \cite{tandon-etal-2018-reasoning}, a seminal dataset of 488 formal science paragraphs, introduced this challenge by asking models to identify changes in entity properties (like location or existence) at each step of a described process (e.g., photosynthesis). OpenPI \cite{tandon-etal-2020-dataset} later extended this idea to 810 real-world how-to guides from WikiHow. It required models to generate structured state-change tuples for each step, capturing transformations. In contrast, FABLE spans a broader range of domains, including fully human-executed procedures (e.g., cooking), semi-automated processes (e.g., travel routes), and fully automated plans (e.g., robotics). It offers an extensible benchmark that evaluates eight distinct forms of data-flow reasoning, encompassing but not limited to entity state tracking.

\begin{table}[t]
\centering
\caption{Comparison of prior procedural reasoning benchmarks with \textsc{FABLE}}
\resizebox{\textwidth}{!}{
\begin{tabular}{@{}p{1.7cm} p{2.5cm} p{2.5cm} p{4.5cm} p{4.5cm}@{}}
\toprule
\textbf{Benchmark} & \textbf{Year (Venue)} & \textbf{Domain} & \textbf{Reasoning Focus} & \textbf{Dataset Size} \\
\midrule
ProPara & 2018 (NAACL) & Science & Entity state tracking (location, existence) & 488 procedures, 81k state annotations\\
WIQA & 2019 (EMNLP) & Science & Causal “what-if” reasoning (perturbation effects) & 379 procedures, 40.7k QA pairs\\
OpenPI & 2020 (EMNLP) & WikiHow guides & Entity state tracking (open vocabulary) & 810 procedures, 29.9k state changes\\
TORQUE & 2020 (EMNLP) & News snippets & Temporal order QA (before/after events) & 3.2k snippets, 21k questions \\
TRACIE & 2021 (NAACL) & News \& stories & Temporal reasoning over implicit events & 1.7k event pairs (implicit vs explicit) \\
CREPE & 2023 (EACL) & Cooking, Household & Causal reasoning with entity state interactions & 183 procedures, 1,219 steps, 324 annotated events \\
\textbf{\textsc{FABLE} (ours)} & \textbf{2025} & \textbf{Recipes, Plans, Travel routes} & \textbf{Data-flow reasoning (state, causal, temporal)} & \textbf{4,260 procedures, 2,400 QA pairs, avg 9.39 steps each} \\
\bottomrule
\end{tabular}
}
% \vspace{-14pt}
\label{tab:procedural-benchmarks}
\end{table}

\begin{figure}[t]
    \centering
    \includegraphics[width=1\linewidth]{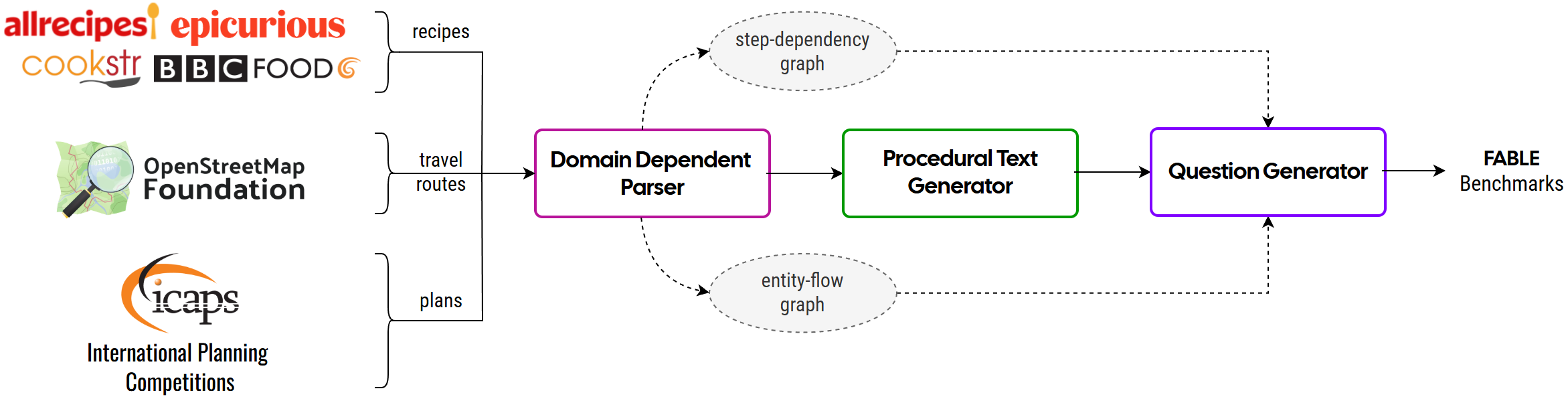}
    \caption{Overview of the FABLE dataset construction pipeline. Raw procedural data is collected from three domains: recipes (human-executed), travel routes (semi-automated), and plans (automated).}
    \label{fig:pipeline}
    % \vspace{-4pt}
\end{figure}

Another line of benchmarks targets causal reasoning in procedural texts, where the goal is to predict or explain the consequences of specific steps within a process. WIQA \cite{tandon2019wiqa} introduced 40,000 counterfactual questions based on 379 procedural texts, each grounded in crowd-sourced influence graphs that capture causal relationships between events. LLMs are evaluated using multiple-choice questions with three answer types: correct effect, opposite effect, and no effect. LLMs are found to struggle with the multi-hop reasoning – tracking chains of influences through the procedure. More recently, CREPE \cite{zhang-etal-2023-causal} was proposed to explicitly link event outcomes with intermediate entity states. This handcrafted benchmark includes 183 procedural texts, such as cooking recipes, where the model is asked, at each step, whether a particular event becomes more likely, less likely, or remains equally likely. For example, after the step ``heat the pan,'' the event ``a sizzling sound occurs'' becomes more likely. While CREPE focuses on single-step causal effects, FABLE goes further by evaluating a model’s understanding of the entire causal structure of a procedure, including branching conditions and interactions. This allows for a deeper and more comprehensive assessment of procedural logic.

Procedural reasoning also requires understanding temporal dependencies, including both the explicit sequence of events and the implicit events that occur in between. Benchmarks such as TORQUE \cite{ning-etal-2020-torque} and TRACIE \cite{zhou-etal-2021-temporal} target temporal reasoning in narrative and historical texts. TORQUE contains 21,000 questions based on 3,200 news snippets, asking what happened before or after a given event. Transformer models often struggle with these tasks; for example, RoBERTa-large achieves only 51\% exact match, approximately 30\% below human performance. TRACIE focuses on implicit event inference, requiring models to identify events not directly mentioned but implied by context—for example, inferring “Alice slept through the night” from “Alice fell asleep. She woke up at dawn.” While these benchmarks help evaluate temporal reasoning, they do so in isolation. In real-world procedures, temporal order, causal effects, and entity states are deeply intertwined. FABLE sets itself apart by integrating these dimensions, requiring models to reason not only about event sequencing but also about how timing impacts outcomes (e.g., “waiting for butter to melt before adding chopped onions”) and how earlier steps affect the feasibility of later ones. Importantly, the majority of these benchmarks have been evaluated using encoder-only models such as BERT and RoBERTa, leaving open the opportunity to assess how well modern autoregressive models handle procedural temporal reasoning in more complex, generative settings.
\section{\textsc{FABLE} Dataset Construction}

In this section, we describe the construction of the FABLE benchmark, starting from domain selection and raw data collection, to parsing, procedural text generation, and the creation of question-answer instances aligned with formal data-flow analyses. An overview of the pipeline is shown in Figure~\ref{fig:pipeline}.

\subsection{Domain Selection and Data Sources}
We select three distinct domains\footnote{We use “domain” to denote application scenarios—recipes, travel routes, and qualify it with IPC when relating to 
%plans—distinct from 
its planning literature usage of problem categories, like Hanoi and Gripper.} for constructing the FABLE benchmark, chosen to span a wide range of procedural complexity and automation levels. The domains include cooking recipes, which are executed by humans, travel routes, which are semi-automated  - followed by either humans or self-driving vehicles, and plans employed in robotics, which are fully automated. This diversity ensures that FABLE evaluates data-flow reasoning across varied forms of procedural text encountered in real-world settings.

\noindent \textbf{Recipes} Cooking recipes represent fully human-executed procedures, rich in causal structure, temporal ordering, and dynamic entity transformations. We draw from the publicly available English Recipes Dataset \cite{Vance_2017}, which compiles recipe data scraped from various online cooking resources. To ensure high-quality procedural structure, we first standardized recipe formats and filtered out instances with fewer than three steps, yielding a base corpus of 74,031 unique recipes across diverse cuisines. To focus on recipes with interpretable and traceable entities, we conducted entity frequency analysis to identify commonly occurring ingredients, tools, and intermediate products. Using spaCy's \texttt{en\_core\_web\_trf} model \cite{Honnibal_spaCy_Industrial-strength_Natural_2020}, we extracted candidate entities and manually curated the list to exclude noisy or uninformative terms such as measurement units and generic descriptors. Recipes containing only entities from this curated vocabulary were retained, resulting in a high-quality subset of 1,382 recipes used for all experiments in this work.

\noindent \textbf{Travel Routes} The travel routes domain consists of structured navigation instructions generated via rule-based systems, representing semi-automated procedures. %followed by either humans or self-driving vehicles. 
We construct this corpus using Valhalla’s open-source turn-by-turn routing API \cite{valhallaopen} which generates natural language driving directions based on OpenStreetMap data. For each start–destination pair, we query Valhalla using five different costing options (e.g., avoid highways, heavily penalize turns) to generate three distinct routes per pair. To ensure procedural validity, we post-process the routes to remove duplicate actions in steps and ensure the routes can be followed as plain text. All locations are sampled from publicly available datasets of airports, historic sites, government buildings, hospitals, and landmarks within the contiguous United States. Each route is constrained to be intra-state and to span a maximum distance of 150 kilometers, ensuring sufficient procedural complexity while maintaining interpretability. The resulting corpus reflects realistic, goal-directed navigation tasks, featuring multi-step, causally ordered, and temporally sequenced instructions.

% Travel routes are textual descriptions generated by rule-based systems to represent a path to be followed by a driver.
% The routes in this benchmark were generated with Valhalla's turn-by-turn route api, using five costing strategies to generate three unique routes for each start and destination. We also made slight modifications to make the routes non-repetitive and consistent. All of the starts and destinations were restricted to be in the same state and no more than 150 km apart. The locations used were chosen at random from public datasets of airports, historic sites, government buildings, hospitals, and landmarks in the contiguous United States. 

\begin{table}[t]
\centering
\caption{Classical data-flow analyses, their procedural text adaptation, and the reasoning type.}
\resizebox{\textwidth}{!}{
\begin{tabular}{@{}p{3.2cm} p{6.0cm} p{6cm} p{2.5cm}@{}}
\toprule
\textbf{Data-Flow Analysis} & \textbf{Software Definition} & \textbf{Procedural Text Adaptation} & \textbf{Reasoning Type} \\
\midrule

\textbf{Reaching Definitions} 
& Tracks which definitions (assignments) of variables can reach a particular program point without being overridden. 
& Tracks whether an entity's production in one step influences its use in subsequent steps.
& State \\

\midrule

\textbf{Very Busy Expressions}
& Identifies expressions that will necessarily be evaluated before any operand changes along all execution paths. 
& Detects entities produced in one step that must be consumed across all future paths.
& Causal \\

\midrule

\textbf{Available Expressions}
& Determines if a previously computed expression is still valid (i.e., not invalidated by redefinition). 
& Verifies if an entity’s computed state can be reused without recomputation in later steps.
& State \\

\midrule

\textbf{Live Variable Analysis}
& Checks whether a variable will be used again before being overwritten or discarded. 
& Determines whether an entity or resource remains necessary for future steps.
& State \\

\midrule

\textbf{Interval Analysis}
& Determines the possible numeric bounds of variables at each program point.
& Tracks valid numeric ranges (e.g., time, quantity) across steps to ensure constraint satisfaction.
& Temporal \\

\midrule

\textbf{Type-State Analysis}
& Ensures objects are used only in valid states (e.g., opening a file before reading it).
& Ensures entities follow valid state transitions across procedural steps.
& State \\

\midrule

\textbf{Taint Analysis}
& Tracks the flow of untrusted or contaminated data to sensitive operations.
& Traces how contaminated or undesirable entities influence procedural steps.
& Causal \\

\midrule

\textbf{Concurrency Analysis}
& Identifies whether operations can be safely executed in parallel without introducing conflicts.
& Detects steps that are independent and can be executed concurrently without dependency violations.
& Temporal \\

\bottomrule
\end{tabular}
}
\label{tab:dataflow-adaptation}
\end{table}

\noindent \textbf{Plans}
Automated planning is the task of generating plans, i.e. a sequence of actions that transforms an initial state into a desired goal state, given a model of the environment \cite{ghallab2004automated}. Formally, a classical planning problem can be defined as a tuple \(\mathcal{P} = (\mathcal{S}, \mathcal{A}, \mathcal{T}, s_0, G)\), where \(\mathcal{S}\) denotes the set of states, \(\mathcal{A}\) denotes the set of actions, \(\mathcal{T}: \mathcal{S} \times \mathcal{A} \rightarrow \mathcal{S}\) is the transition function that defines the effect of actions, \(s_0 \in \mathcal{S}\) is the initial state, and \(G \subseteq \mathcal{S}\) is the set of goal states \cite{russell2005ai}. The International Planning Competition (IPC) \cite{ipc}, organized as part of the premier planning conference \textit{ICAPS}, releases standardized planning domains and problem generators to benchmark automated planners. We use four classical IPC domains in our dataset construction: Hanoi, Gripper, Ferry, and Driverlog \cite{pddl-generators}. Each domain introduces distinct challenges, such as sequential dependencies, resource transportation, and object manipulation. We employ the Fast Downward planning system \cite{helmert2006fast} to automatically generate ground-truth plans from domain and problem specifications. In total, we generate 1,378 distinct plans across these domains. These plans serve as the procedural basis for generating structured texts and question-answer instances in the \textsc{FABLE} benchmark.

\subsection{Domain Dependent Parser}
\label{sec:parser}

Given raw data from each domain, the first step in the FABLE construction pipeline is to convert unstructured procedural information into structured representations that expose the underlying data-flow and step dependencies. This transformation is handled by a \textit{Domain Dependent Parser}, customized to the characteristics of each domain: recipes, travel routes, and automated plans. For each input, the parser produces two outputs: a \textbf{Step-Dependency Graph} \(G_S = (S, E_S)\), where \(S\) is the set of steps and \(E_S \subseteq S \times S\) encodes ordering constraints between steps; and an \textbf{Entity-Flow Graph} \(G_E = (V, E_E)\), where \(V\) is the set of entities or variables and \(E_E \subseteq S \times V \times S\) captures the flow of entities through the procedure, including creation, modification, and consumption events.

Parsing strategies are tailored to the domain structure. For recipes, we apply natural language processing techniques to identify verbs (actions) and their arguments (ingredients, tools), using temporal markers such as “after stirring” or “meanwhile” to infer step dependencies and tracking ingredient transformations as evolving entity states. For travel routes, structured metadata from OpenStreetMap is leveraged to extract waypoints, transportation modes, and transfer dependencies, with step order implicitly determined by sequential travel progressions.For automated plans, grounded STRIPS-style action schemas define preconditions and effects \cite{mcdermott1998pddl}, enabling the direct construction of a step-dependency graph. In parallel, we construct an entity-flow graph by tracking how predicates and object attributes are modified across actions, capturing how entities evolve throughout the plan execution.

\subsection{Procedural Text Generator}

The next step in the FABLE construction pipeline is to generate coherent natural language procedural texts. The goal of this stage is to unify all domains under a common linguistic format, while preserving the underlying structure of goals, steps, and ordering constraints. Procedures originating from recipes are already provided in natural language, containing an explicit goal \(G\) describing the final desired dish, and a set of steps \(S = \{s_1, \dots, s_n\}\) corresponding to individual cooking instructions. Temporal and causal markers present in the recipe text, such as ``meanwhile'' or ``after mixing,'' are used to infer the partial ordering \(\preceq\) over steps. The parsing process identifies and extracts \(G\), \(S\), and \(\preceq\) to construct a structured procedural representation without altering the inherent ordering information conveyed in the original text. For travel routes, the goal \(G\) is defined by a source and destination pair, specified as coordinate locations. The set of steps \(S = \{s_1, \dots, s_n\}\) consists of human-readable navigation instructions extracted directly from structured route metadata provided by OpenStreetMap, such as ``Turn left onto Airport Boulevard'' or ``Take the US 10 East ramp toward Bay City.'' The ordering \(\preceq\) in this domain corresponds to a total order, where each step \(s_i\) must be executed immediately before \(s_{i+1}\), reflecting the sequential nature of travel from the source to the destination. For automated plans, the raw data consists of grounded action sequences specified in PDDL (Planning Domain Definition Language). The goal \(G\) corresponds to the desired final state specified in the planning problem. The set of steps \(S = \{s_1, \dots, s_n\}\) is derived from the sequence of grounded actions that form a valid plan achieving \(G\) from the initial state. To convert these symbolic actions into natural language, we use AutoPlanBench \cite{stein2023autoplanbench}, a planning-to-text system that maps each grounded action into a fluent English instruction while preserving its semantic role in the overall plan, which follows a total order similar to travel routes. By standardizing representation across diverse domains, FABLE enables a unified evaluation framework for data-flow reasoning over procedural text.

\begin{table}[t]
\centering
\caption{Summary of dataset statistics across the three domains in \textsc{FABLE}.}
\resizebox{\textwidth}{!}{
\begin{tabular}{@{}lcccccc@{}}
\toprule
\textbf{Domain} 
& \makecell{\textbf{Total}\\\textbf{Procedures}} 
& \makecell{\textbf{Avg. Steps}\\\textbf{per Procedure}} 
& \makecell{\textbf{Min}\\\textbf{Steps}} 
& \makecell{\textbf{Max}\\\textbf{Steps}} 
& \makecell{\textbf{Total QA Pairs}\\\textbf{Generated}} 
& \makecell{\textbf{QA Pairs Used}\\\textbf{in \textsc{FABLE}}} \\
\midrule

Recipes & 1,382 & 4.15 & 3 & 17 & 1,600 & 800 \\

Travel Routes & 1,500 & 17.26 & 4 & 56 & 800 & 800 \\

Plans & 1,378 & 6.77 & 1 & 15 & 11,024 & 800 \\

\midrule
\textbf{Total} & 4,260 & 9.39 & - & - & 13,424 & 2,400 \\

\bottomrule
\end{tabular}
}
\label{tab:fable-statistics}
\end{table}

\subsection{Question Generator}
\label{sec:question-generator}

The final stage of FABLE construction involves generating diagnostic question-answer instances that probe a model's data-flow reasoning abilities over procedural texts. This is achieved by systematically adapting classical data-flow analyses from software engineering and framing them as natural language questions grounded in the structure of each procedure. Table~\ref{tab:dataflow-adaptation} summarizes the original definitions of classical data-flow analyses, their adaptation for procedural texts, and the type of reasoning they primarily evaluate. Given the step-dependency graph \(G_S = (S, E_S)\) and the entity-flow graph \(G_E = (V, E_E)\) extracted during parsing, we employ a template-based generation approach to create question instances. For each data-flow analysis, we define canonical templates that map properties of \(G_S\) and \(G_E\) to corresponding natural language questions and formally determined answers. This allows for scalable and consistent generation across domains and procedures without requiring manual annotation. For all analyses except Interval Analysis, the generated questions are structured as binary \textit{yes/no} queries, directly answerable from the properties of \(G_S\) and \(G_E\). For Interval Analysis, the answers involve either a numerical range (e.g., "the oven temperature is maintained between 350°F and 375°F") or identification of specific steps within \(S\) where a numeric attribute condition is satisfied (e.g., "between Step~4 and Step~7"). Certain analyses yield deterministic answers in specific domains due to structural regularities in the underlying graphs. In travel routes, questions on Available Expressions and Taint Analysis always evaluate to \textit{Yes}, as navigation steps monotonically accumulate accessible locations and no semantic contamination of route entities occurs. Conversely, Concurrency, Type-State, and Very Busy Expressions consistently yield \textit{No} due to the strictly sequential nature of route instructions, which lack side-effect reuse or conditional branching in \(G_E\) and \(G_S\). Similarly, in recipes, Reaching Definitions and Available Expressions always evaluate to \textit{Yes}, reflecting the persistent availability of ingredients once introduced. Type-State and Concurrency Analyses consistently yield \textit{No}, as most actions have strict temporal dependencies and invalid skipping of preparatory steps. Taint Analysis returns \textit{Yes}, as cooking inherently sanitizes inputs. These outcomes are a direct consequence of deterministic semantics in entity flow and step ordering within these procedural domains, not of annotation bias or trivial task construction. Table~\ref{tab:fable-statistics} summarizes the composition of the \textsc{FABLE} benchmark across domains, including the number of procedures, the distribution of procedural lengths (in steps), and the number of question-answer pairs generated via data-flow analyses. Although over 13,000 QA instances were automatically generated, a balanced subset of 800 per domain (2,400 total) was selected to ensure coverage and consistency across analyses and domains.
\section{Experimental Setup \& Results}

The goal of our experimental evaluation is to investigate the following research questions:
\begin{enumerate*}[label=\textbf{(\arabic*)}]
    \item Is \textsc{FABLE} a challenging benchmark for current large language models?
    \item What is the baseline performance across different types of LLM models?
\end{enumerate*} To answer these questions, we systematically evaluate representative large language models on the FABLE benchmark and analyze their performance across data-flow analyses and domains. We first describe the benchmark models used in our experiments, followed by the evaluation setup and results corresponding to each research question.

\noindent \textbf{Baseline Models}

To benchmark the performance of LLMs on \textsc{FABLE} and assess the difficulty of its procedural reasoning tasks, we evaluate three representative categories of models: a reasoning-focused model (\texttt{deepseek-r1:8b}), a general-purpose model (\texttt{llama3.1:8b}), and a code-specific model (\texttt{granite-code:8b}). These categories are chosen to capture different design priorities: abstract multi-step reasoning, broad natural language competence, and structured procedural understanding, respectively. To ensure a fair comparison, all models are approximately 8 billion parameters in size and are accessed through the Ollama platform under consistent inference settings. Table~\ref{tab:benchmark-models} summarizes the key characteristics of the models used in our experiments.

\begin{table}[t]
\centering
\caption{Benchmark models evaluated on FABLE.}
\resizebox{\textwidth}{!}{
\begin{tabular}{@{}lcccccc@{}}
\toprule
\textbf{Model} & \textbf{Architecture} & \textbf{Parameters} & \textbf{Quantization} & \textbf{Model Size} & \textbf{License} & \textbf{Source} \\
\midrule
\texttt{deepseek-r1:8b} & LLaMA & 8.03B & Q4\_K\_M & 4.9 GB & MIT License & \includegraphics[height=0.4cm]{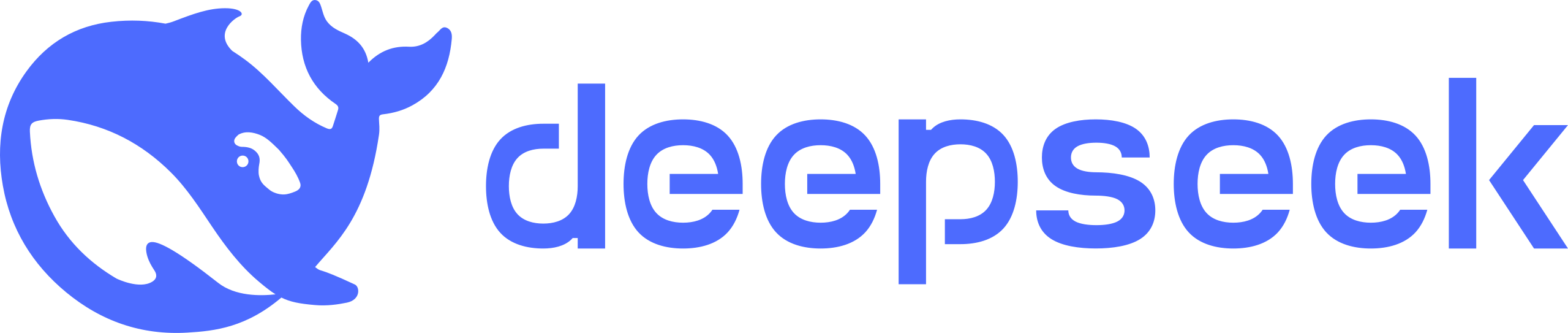} \\
\texttt{llama3.1:8b} & LLaMA & 8.03B & Q4\_K\_M & 4.9 GB & Llama 3.1 Community License & \includegraphics[height=0.25cm]{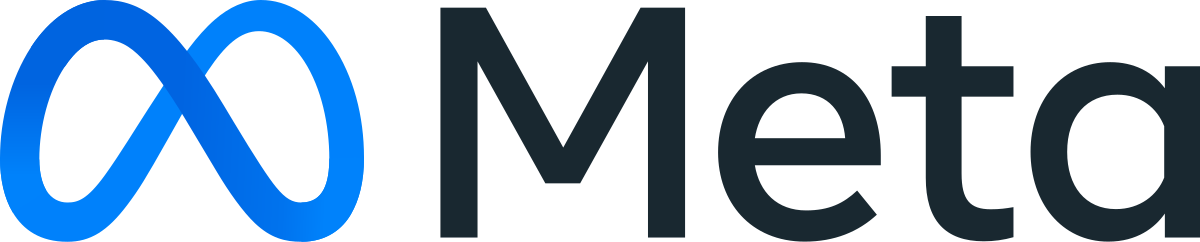} \\
\texttt{granite-code:8b} & LLaMA & 8.05B & Q4\_0 & 4.6 GB & Apache 2.0 & \includegraphics[height=0.27cm]{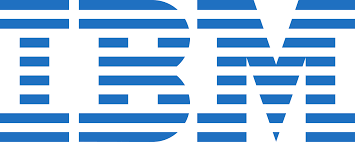} \\
\bottomrule
\end{tabular}
}
\label{tab:benchmark-models}
\end{table}

\noindent \textbf{Evaluation Setup}

Each model is evaluated across the full FABLE benchmark, covering all domains and data-flow analyses. For every question instance, we generate \textbf{five independent completions} using greedy decoding, with sampling seeds randomized across trials kept constant across experiments. Final predictions are determined through \textbf{majority voting} over the sampled completions to reduce stochastic variability and improve answer robustness. For binary (yes/no) questions, we report accuracy as the proportion of correct predictions compared to ground-truth annotations. For interval-based questions, a prediction is considered correct if the extracted value lies within the expected numerical range or correctly identifies the relevant step interval as specified by the procedural structure. We use the Ollama Python API \cite{marcondes2025using} to run inference over each prompt, using the chat-based interface with a single-user message and streamed decoding enabled. Each model processes the prompts sequentially, and inference time is recorded per model–analysis pair to assess runtime cost. Responses are collected along with the ground truth and prompt for post-hoc evaluation. Unless otherwise specified, models are run using Ollama’s default decoding parameters, with no manual constraints on maximum generation length, temperature, or sampling strategy. No batching or caching is used during evaluation. Each prompt is processed independently to ensure consistent timing and sampling behavior across trials. All evaluations are conducted locally using quantized 4-bit models served through the Ollama platform. Inference is performed on a single machine equipped with an AMD Ryzen 7 7800X3D 8-core processor, 32~GB RAM, and an NVIDIA GeForce RTX 4080 SUPER GPU with 16~GB VRAM. Each model is loaded and served independently under identical settings.

\noindent \textbf{Results and Analysis}

We present the results of our evaluation organized around the two research questions introduced earlier. We first examine whether FABLE poses significant challenges for the baseline models, followed by a comparison of performance across model categories.

\textbf{\textit{Is \textsc{FABLE} a challenging benchmark for current LLMs?}}

\begin{wrapfigure}{r}{0.65\linewidth}
    \centering
    \includegraphics[width=\linewidth]{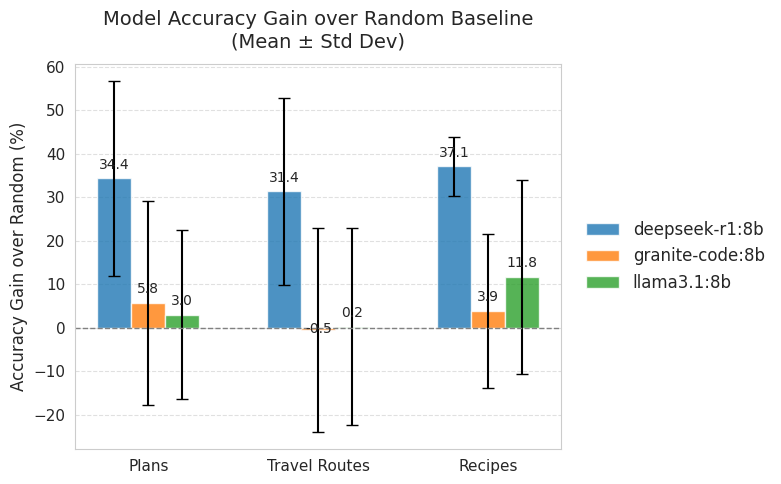}
    \caption{Domain-wise accuracy gains over random baseline for each model, averaged across eight data-flow analyses. Error bars show standard deviation. \texttt{deepseek-r1:8b} shows consistent, significant improvements, while others show marginal or negative gains. These results highlight the procedural reasoning challenges posed by \textsc{FABLE}.}
    \label{fig:delta-over-random}
\end{wrapfigure}
To assess whether \textsc{FABLE} presents a substantive challenge for LLMs, we compare their performance against structured random baselines tailored to each domain and data-flow analysis. These baselines are constructed using template-specific sampling strategies for each analysis. For binary tasks, responses are drawn uniformly between ``Yes'' and ``No''. For interval analyses, domain-specific value ranges are inferred from ground truth annotations, and random intervals are generated accordingly—either numeric ranges (e.g., time or temperature bounds) or step-indexed intervals (e.g., ``Between Step~4 and Step~7'') depending on the domain. Figure~\ref{fig:delta-over-random} plots, for each model and domain, the amount by which model accuracy exceeds its random–guess baseline (i.e., model accuracy minus random baseline accuracy).

% \begin{figure}
%     \centering
%     \includegraphics[width=0.6\linewidth]{figures/delta_over_random1.png}
%     \caption{Domain-wise delta in accuracy over random baseline for each model, aggregated across eight data-flow analyses. Each bar represents the average performance improvement over the random baseline in a given domain, with error bars indicating standard deviation across the eight analyses. While \texttt{deepseek-r1:8b} demonstrates consistent and substantial gains, the other models exhibit marginal improvements or even underperform relative to random in certain domains. These results highlight the procedural reasoning challenges posed by \textsc{FABLE}.
% }
%     \label{fig:delta-over-random}
% \end{figure}

\texttt{deepseek-r1:8b} demonstrates significant improvement over random guessing across all three domains, surpassing the baseline by 34.4\% in plans, 31.4\% in travel routes, and 37.1\% in recipes. In contrast, both \texttt{granite-code:8b} and \texttt{llama3.1:8b} yield only marginal gains: \texttt{granite-code:8b} achieves a modest 5.8\% improvement in plans, fails to outperform random guessing in travel routes (\textit{–0.5\%}), and shows just 3.9\% improvement in recipes. Similarly, \texttt{llama3.1:8b} records minor deltas of 3.0\%, 0.2\%, and 11.8\% across the three domains, respectively. Crucially, these values cluster near the performance floor defined by random selection—underscoring that without specialized reasoning capabilities, general-purpose and code-centric LLMs are effectively guessing on \textsc{FABLE}’s procedural tasks. Furthermore, the large standard deviations observed—frequently exceeding ±20 percentage points indicate that even \texttt{deepseek-r1:8b}'s advantage is unevenly distributed across analyses. This variability highlights the non-uniform difficulty of \textsc{FABLE}’s constituent reasoning types, particularly those involving state persistence, temporal constraints, and causal chains. {\bf These findings reaffirm that \textsc{FABLE} presents a rigorous diagnostic challenge}: robust performance demands more than surface-level fluency, requiring deliberate reasoning architectures that can systematically track and manipulate evolving procedural structures.

\textbf{\textit{What is the baseline data flow reasoning performance across different model classes?}}

\begin{figure}[htbp]
    \centering
    \includegraphics[width=1\linewidth]{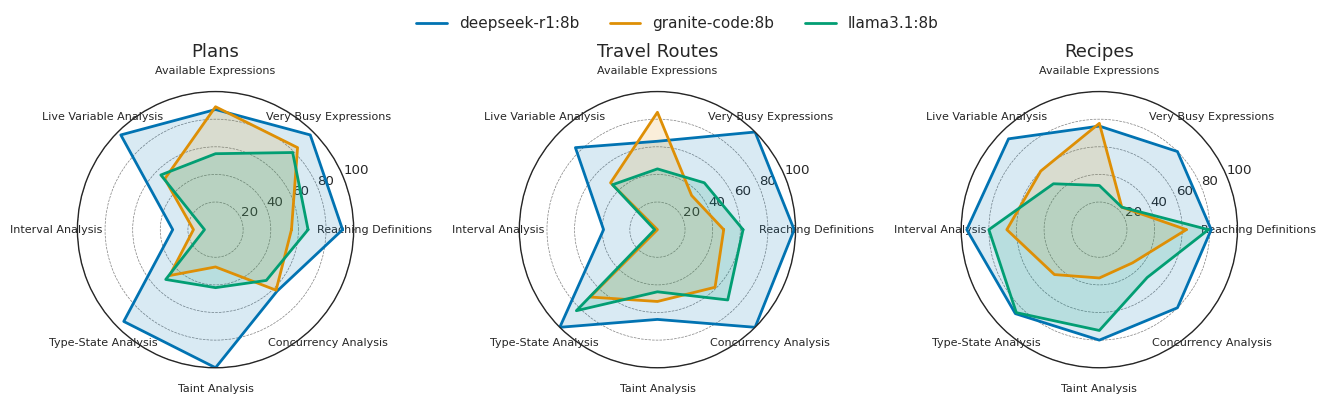}
    \caption{Model-wise performance across domains and data-flow analyses in \textsc{FABLE}}
    \label{fig:radar-all}
\end{figure}

We evaluate three representative model classes on \textsc{FABLE} to establish strong procedural reasoning baselines: \texttt{deepseek-r1:8b}, \texttt{granite-code:8b}, and \texttt{llama3.1:8b}. Figure~\ref{fig:radar-all} presents a comparative view of model performance across eight data-flow analyses within three domains on a shared radial scale. \texttt{deepseek-r1:8b} shows consistently higher scores, forming broader and more stable profiles, while \texttt{granite-code:8b} and \texttt{llama3.1:8b} exhibit irregular shapes with sharp performance drops in several analyses.

The most prominent failure mode across models and domains lies in Interval Analysis. Tasks in this category require interpreting or inferring numeric ranges based on procedural text. In the plans and travel routes, \texttt{granite-code:8b} and \texttt{llama3.1:8b} perform near chance level (0–16\%), and even the reasoning-focused \texttt{deepseek-r1:8b} achieves only 31\% and 39\% in travel routes and plans respectively. These intervals typically involve short but tightly structured sequences with implicit dependencies, making range inference a compositional task that challenges the pattern-matching tendencies of LLMs. In contrast, the same models perform better on Interval Analysis in recipes domain likely reflecting the prevalence of recipes in pretraining corpora and the syntactic familiarity of instructions involving amounts, durations, or temperatures enabling more effective template matching without deep inference. The domain-specific nature of success in Interval Analysis exposes a critical weakness: current LLMs rely heavily on surface-form alignment with training data rather than abstract numerical reasoning. Beyond Interval Analysis, a broader pattern emerges: model class significantly influences performance. \texttt{deepseek-r1:8b} achieves the highest average accuracy across all domains (above 80\%), but with standard deviations around 25\%, indicating substantial variability across data-flow analyses. \texttt{granite-code:8b} and \texttt{llama3.1:8b}, by contrast, exhibit lower accuracy and greater volatility, often failing on analyses that require implicit state tracking or temporally grounded inferences (e.g., Taint or Concurrency Analysis). Notably, \texttt{granite-code:8b} outperforms \texttt{llama3.1:8b} on certain syntactic tasks like Available and Very Busy Expressions, possibly due to its exposure to control-flow patterns during code pretraining. However, it collapses on more dynamic analyses, indicating that local symbolic regularities are not sufficient for robust procedural understanding. \texttt{llama3.1:8b}, though more general, shows modest resilience on certain temporal tasks in recipes, but lacks the structure-aware capabilities required for systematic generalization. {\bf Overall, \textsc{FABLE} offers significant challenges to models in data flow analyses, which varies with model type but is hardest for interval and variable analyses}.

\begin{table}[ht]
\centering
\caption{Average inference time (in seconds) per 100 prompts across domains, with slowdown computed relative to the fastest model in each domain.}
\label{tab:inference-latency}
\begin{tabular}{lcccccc}
\toprule
\textbf{Model} & \multicolumn{2}{c}{\textbf{Plans}} & \multicolumn{2}{c}{\textbf{Travel Routes}} & \multicolumn{2}{c}{\textbf{Recipes}} \\
\cmidrule(lr){2-3} \cmidrule(lr){4-5} \cmidrule(lr){6-7}
& Time (s) & Slowdown & Time (s) & Slowdown & Time (s) & Slowdown \\
\midrule
\texttt{deepseek-r1:8b}   & 1136.3 & 29.6$\times$ & 1059.6 & 37.6$\times$ & 604.4 & 21.4$\times$ \\
\texttt{granite-code:8b}  & 37.5   & 1.0$\times$  & 46.2   & 1.6$\times$  & 46.2  & 1.6$\times$  \\
\texttt{llama3.1:8b}      & 38.3   & 1.0$\times$  & 28.2   & 1.0$\times$  & 28.2  & 1.0$\times$  \\
\bottomrule
\end{tabular}
\end{table}

Table~\ref{tab:inference-latency} reports the average inference time (in seconds) per 100 prompts for each model across the three domains in \textsc{FABLE}, along with the corresponding slowdown relative to the fastest model in that domain. The slowdown is defined as \(\text{Slowdown}_{A:B} = T_A / T_B\), where \(T_A\) and \(T_B\) denote the inference times of models \(A\) and \(B\), respectively. A slowdown of \(x\times\) indicates that model \(A\) is \(x\) times slower than the fastest model \(B\) on that domain. While \texttt{deepseek-r1:8b} demonstrates superior accuracy, it incurs substantially higher inference cost—over \(20\times\) slower than the fastest model in each domain. {\bf This pronounced disparity highlights a critical trade-off between model performance and deployment efficiency, particularly relevant for real-time or resource-constrained applications}.

\section{Conclusion}

This work introduced \textsc{FABLE}, a novel benchmark designed to evaluate the data-flow reasoning capabilities of LLMs over procedural text. By adapting classical data-flow analyses from software engineering into natural language tasks, we create a rigorous and interpretable diagnostic suite spanning automated plans, travel routes, and recipes. Our evaluation reveals that while reasoning-focused models like \texttt{deepseek-r1:8b} can outperform general-purpose and code-specific models, all models exhibit significant variance across reasoning types and struggle especially with numeric and temporal analyses like Interval and Concurrency. The structured nature of \textsc{FABLE} enables fine-grained evaluation of LLMs along dimensions often overlooked in existing benchmarks, including compositional inference, causal ordering, and entity-state tracking. Furthermore, our results expose a clear gap between surface-level success and robust procedural understanding, motivating future work on hybrid neurosymbolic methods, structured fine-tuning, and targeted pretraining. A detailed discussion of limitations and broader impacts is provided in the Appendix. 

\bibliographystyle{plain}
\bibliography{references}

%%%%%%%%%%%%%%%%%%%%%%%%%%%%%%%%%%%%%%%%%%%%%%%%%%%%%%%%%%%%
%%%%%%%%%%%%%%%%%%%%%%%%%%%%%%%%%%%%%%%%%%%%%%%%%%%%%%%%%%%%

\section*{NeurIPS Paper Checklist}

\begin{enumerate}

\item {\bf Claims}
    \item[] Question: Do the main claims made in the abstract and introduction accurately reflect the paper's contributions and scope?
    \item[] Answer: \answerYes{}.
    \item[] Justification: The main claims have been made in the abstract and also in the introduction and addressed in Section 4 pages 8-9.

\item {\bf Limitations}
    \item[] Question: Does the paper discuss the limitations of the work performed by the authors?
    \item[] Answer: \answerYes{} 
    \item[] Justification: The limitations regarding the data-flow analyses adapted from software engineering to procedural text and the experiments are mentioned in Appendix A.1.

\item {\bf Theory assumptions and proofs}
    \item[] Question: For each theoretical result, does the paper provide the full set of assumptions and a complete (and correct) proof?
    \item[] Answer: \answerNA{} % Replace by \answerYes{}, \answerNo{}, or \answerNA{}.
    \item[] Justification: There are no theoretical proofs in this paper.

    \item {\bf Experimental result reproducibility}
    \item[] Question: Does the paper fully disclose all the information needed to reproduce the main experimental results of the paper to the extent that it affects the main claims and/or conclusions of the paper (regardless of whether the code and data are provided or not)?
    \item[] Answer: \answerYes{} 
    \item[] Justification: All the experimental results and raw data used for creating the \textsc{FABLE} dataset and benchmarks are uploaded on Github repository for reproducibility - \url{https://github.com/VishalPallagani/FABLE/}

\item {\bf Open access to data and code}
    \item[] Question: Does the paper provide open access to the data and code, with sufficient instructions to faithfully reproduce the main experimental results, as described in supplemental material?
    \item[] Answer: \answerYes{}
    \item[] Justification: The Github repository consists of the data and code - \url{https://github.com/VishalPallagani/FABLE/} and also data is available as a Hugging Face dataset at \url{https://huggingface.co/datasets/g-nitin/FABLE}

\item {\bf Experimental setting/details}
    \item[] Question: Does the paper specify all the training and test details (e.g., data splits, hyperparameters, how they were chosen, type of optimizer, etc.) necessary to understand the results?
    \item[] Answer: \answerYes{} % Replace by \answerYes{}, \answerNo{}, or \answerNA{}.
    \item[] Justification: The test settings are mentioned in Evaluation Setup in Section 4, page 7.

\item {\bf Experiment statistical significance}
    \item[] Question: Does the paper report error bars suitably and correctly defined or other appropriate information about the statistical significance of the experiments?
    \item[] Answer: \answerYes{} % Replace by \answerYes{}, \answerNo{}, or \answerNA{}.
    \item[] Justification: Error bars are represented in Figure 2 of the main paper and more details in the Appendix.

\item {\bf Experiments compute resources}
    \item[] Question: For each experiment, does the paper provide sufficient information on the computer resources (type of compute workers, memory, time of execution) needed to reproduce the experiments?
    \item[] Answer: \answerYes{} % Replace by \answerYes{}, \answerNo{}, or \answerNA{}.
    \item[] Justification: The Evaluation Setup in Section 4, page 7 mentions the compute resources used for the experiments.
    
\item {\bf Code of ethics}
    \item[] Question: Does the research conducted in the paper conform, in every respect, with the NeurIPS Code of Ethics \url{https://neurips.cc/public/EthicsGuidelines}?
    \item[] Answer: \answerYes{} % Replace by \answerYes{}, \answerNo{}, or \answerNA{}.
    \item[] Justification: The papers conforms to the NeurIPS Code of Ethics.

\item {\bf Broader impacts}
    \item[] Question: Does the paper discuss both potential positive societal impacts and negative societal impacts of the work performed?
    \item[] Answer: \answerYes{} % Replace by \answerYes{}, \answerNo{}, or \answerNA{}.
    \item[] Justification: The broader impacts of the work are presented in Appendix A.2.
    
\item {\bf Safeguards}
    \item[] Question: Does the paper describe safeguards that have been put in place for responsible release of data or models that have a high risk for misuse (e.g., pretrained language models, image generators, or scraped datasets)?
    \item[] Answer: \answerNA{} % Replace by \answerYes{}, \answerNo{}, or \answerNA{}.
    \item[] Justification: The paper poses no such risks as we use open data to prepare reasoning benchmarks.

\item {\bf Licenses for existing assets}
    \item[] Question: Are the creators or original owners of assets (e.g., code, data, models), used in the paper, properly credited and are the license and terms of use explicitly mentioned and properly respected?
    \item[] Answer: \answerYes{} % Replace by \answerYes{}, \answerNo{}, or \answerNA{}.
    \item[] Justification: Yes, all data used in the paper is open data.

\item {\bf New assets}
    \item[] Question: Are new assets introduced in the paper well documented and is the documentation provided alongside the assets?
    \item[] Answer: \answerYes{} % Replace by \answerYes{}, \answerNo{}, or \answerNA{}.
    \item[] Justification: The new dataset and benchmarks introduced are well documented both as a dataset on Hugging Face and also on Github.

\item {\bf Crowdsourcing and research with human subjects}
    \item[] Question: For crowdsourcing experiments and research with human subjects, does the paper include the full text of instructions given to participants and screenshots, if applicable, as well as details about compensation (if any)? 
    \item[] Answer: \answerNA{} % Replace by \answerYes{}, \answerNo{}, or \answerNA{}.
    \item[] Justification: The paper does not involve crowdsourcing or research with humans.

\item {\bf Institutional review board (IRB) approvals or equivalent for research with human subjects}
    \item[] Question: Does the paper describe potential risks incurred by study participants, whether such risks were disclosed to the subjects, and whether Institutional Review Board (IRB) approvals (or an equivalent approval/review based on the requirements of your country or institution) were obtained?
    \item[] Answer: \answerNA{} % Replace by \answerYes{}, \answerNo{}, or \answerNA{}.
    \item[] Justification: The paper does not involve crowdsourcing nor research with human subjects.

\item {\bf Declaration of LLM usage}
    \item[] Question: Does the paper describe the usage of LLMs if it is an important, original, or non-standard component of the core methods in this research? Note that if the LLM is used only for writing, editing, or formatting purposes and does not impact the core methodology, scientific rigorousness, or originality of the research, declaration is not required.
    %this research? 
    \item[] Answer: \answerYes{} % Replace by \answerYes{}, \answerNo{}, or \answerNA{}.
    \item[] Justification: The LLM usage has been declared in Appendix A.3 and also in the OpenReview submission page.

\end{enumerate}

\newpage
\appendix

\section{Technical Appendices and Supplementary Material}

% \begingroup
% \renewcommand{\contentsname}{Appendix Contents}
% \setcounter{tocdepth}{2}
% \tableofcontents
% \endgroup

\section*{Appendix Contents}

\begin{flushleft}
\begin{tabular}{@{}ll}
\quad A.1 \hspace{1em} Limitations \& Future Work \dotfill & 15 \\
\quad A.2 \hspace{1em} Broader Impacts \dotfill & 16 \\
\quad A.3 \hspace{1em} More Information on Data Sources \dotfill & 16 \\
\quad A.4 \hspace{1em} Prompt Examples for Data-Flow Analyses across Domains \dotfill & 18 \\
\quad A.5 \hspace{1em} Domain-wise Results and Statistical Significance \dotfill & 31 \\
\quad A.6 \hspace{1em} Inference Time Analysis \dotfill & 32 \\
\end{tabular}
\end{flushleft}

\subsection{Limitations \& Future Work}
While \textsc{FABLE} offers a principled benchmark for evaluating data-flow reasoning in LLMs, some limitations remain. First, the dataset is constructed over three manually curated domains—recipes, travel routes, and automated plans—which introduces domain-specific biases. Automated plans, due to their formal structure, are easier to parse and verify, while human-authored texts like recipes require manual curation and domain-specific heuristics, limiting scalability and generalization to other procedural domains. Second, the adaptation of classical data-flow analyses to procedural text reflects only one formal interpretation per analysis. Alternative definitions or relaxed variants could yield different insights, but \textsc{FABLE} currently encodes only a single adaptation for each case. Third, model evaluation is limited to a fixed prompting style using template-based instructions. Variations in prompt phrasing, such as chain-of-thought or few-shot prompting, could significantly impact model performance and reveal hidden capabilities or weaknesses. Finally, the reliance on binary question-answer formats may obscure intermediate reasoning errors or partial correctness. Richer response formats, such as justifications or step-level annotations, could provide deeper insight into reasoning trajectories and failure modes.

Future work will focus on expanding both the breadth and depth of the benchmark. On the breadth side, we plan to incorporate additional domains such as conversational dialogs, design diagrams, and scientific protocols, which would allow for a more comprehensive assessment of procedural reasoning across settings with varied linguistic and structural properties. On the depth side, we aim to scale the number of QA instances per domain and analysis type to enable finer-grained model comparisons and robust statistical evaluations. We also envision extending the benchmark to support multiple data-flow formalizations for each analysis, enabling a more nuanced understanding of how different reasoning frameworks interact with model behavior. Beyond dataset augmentation, future work will investigate more expressive evaluation schemes, including free-form rationales, multi-step explanations, and interaction-based formats that capture how LLMs revise their reasoning upon receiving intermediate feedback. Finally, integrating semi-automated verification tools and feedback-driven prompting strategies could further support benchmarking of emergent LLM capabilities under realistic usage conditions.

\subsection{Broader Impacts}
\textsc{FABLE} benchmark has the potential to advance both foundational and applied research in procedural understanding, reasoning diagnostics, and systematic model evaluation. By grounding classical data-flow analyses in natural language, \textsc{FABLE} bridges concepts from software verification and natural language processing, enabling a new direction for evaluating compositional reasoning, entity tracking, and procedural consistency in LLMs.

From an applied perspective, this work can inform the development of safer and more reliable AI systems for domains where procedural accuracy is critical, such as healthcare (e.g., medical protocols), robotics (e.g., task planning), and education (e.g., instructional feedback). At the same time, the benchmark highlights current model deficiencies in handling temporal, causal, and state-based dependencies, potentially guiding the design of more interpretable and trustworthy AI systems.

There are minimal risks of misuse, as the dataset is constructed entirely from publicly available or procedurally generated content and contains no sensitive or private data. Nonetheless, improvements in procedural reasoning could amplify the deployment of LLMs in high-stakes environments. Therefore, it is important that future applications of such models incorporate safeguards such as verification modules, human-in-the-loop checks, or counterfactual simulations to mitigate the impact of incorrect inferences. We hope that \textsc{FABLE} contributes to more deliberate and transparent evaluation practices in the growing ecosystem of language model benchmarking.

\subsection{More Information on Data Sources}
\subsubsection{Recipes}

To procure 800 recipe QA pairs for the FABLE benchmark, we begin with the English Recipes Dataset \cite{Vance_2017}, a publicly released English-language recipe archive collected in June–July 2017. This snapshot aggregates data from four major culinary websites: \textit{Allrecipes.com} ($\approx 91,000$ recipes), \textit{Epicurious.com} ($\approx 34,000$), \textit{BBC.co.uk}’s Food section ($\approx 10,000$), and \textit{Cookstr.com} ($\approx 8,000$). Each website's data included photographs, raw HTML or JSON payloads, and the scripts used to crawl them. However, since our downstream parsing and question-generation pipeline requires fully structured ingredient lists and coherent step sequences, we elected to restrict our analysis to the Allrecipes.com subset and re-crawled every URL supplied in the subset. This resulted in recipe records that included, among other irrelevant components, a title, an (optional) ingredient list, and a free-form `instructions list' of natural language steps. To guarantee a minimally informative procedural structure, we discarded any recipe whose instructions contained fewer than three steps, yielding $74,031$ recipes. Each instruction was treated as a discrete step $s_i$, preserving its original ordering and any temporal or causal connectives (e.g.\ `meanwhile', `after').

We then applied our domain‐dependent parser (see Section~\ref{sec:parser}) with spaCy’s \texttt{en\_core\_web\_trf} model~\cite{Honnibal_spaCy_Industrial-strength_Natural_2020} to extract candidate entities from each step. Noun‐chunk extraction was followed by a sequence of rule‐based filters: removal of measurement units (e.g.\ `cup', `tsp'), container and tool names when uninformative (e.g.\ `bowl', `spoon'), generic descriptors (e.g.\ `mixture', `amount'), numeric or dimensional patterns (e.g.\ `9×13', `10–15'), and verb‐derived phrases.  We also applied derived‐entity heuristics to link suffix forms (e.g.\ `batter', `sauce') back to their base entity when appropriate.

We performed a frequency analysis on this initial entity pool to focus our question generation machinery on the most salient domain concepts. We ranked entities by occurrence count and iteratively selected the smallest subset that, under a simple text‐mention heuristic, would yield at least 100 questions per data‐flow category. A subsequent manual review, consisting of removing residual noisy terms (e.g.\ `ingredients', `rest') and overly generic labels, produced a final vocabulary of domain‐relevant entities, encompassing core ingredients (e.g.\ `eggs', `flour'), intermediate preparations (e.g.\ `dough', `glaze') and essential tools or appliances (e.g.\ `oven', `mixer').

Finally, we filtered the $ 74,031$-recipe corpus to retain only those recipes whose extracted entities lie entirely within our curated vocabulary.  This ensures that, for every retained recipe, all entities referenced in questions and answers can be unambiguously grounded in the underlying text.  The resulting high‐quality subset consists of $1,382$ recipes, which serve as the basis for all subsequent question‐answer generation in the recipes domain.

From the curated set of $1,382$ recipes, we instantiated our eight data‐flow analysis templates on each procedure, producing an initial, larger pool of $1,600$ question–answer instances. We then applied a strict relevance filter, retaining only those questions that explicitly mention at least one entity from our manually vetted vocabulary. The remaining candidates were grouped by analysis type and sorted to maximize diversity of source recipes; from each group, we selected exactly 100 questions, yielding a balanced set of 800 recipe QA pairs.

Each of the 800 selected question–answer pairs underwent a final two‐phase curation. In the first phase, two co-authors reviewed every answer for correctness, marking any that appeared inconsistent with the underlying procedural representation. In the second phase, for each flagged item, we algorithmically generated alternative candidate questions that were re-verified until the 100‐per‐type quota was restored. This ensured that every analysis category in the recipes domain consists of exactly 100 fully validated QA instances.

\subsubsection{Travel Routes}

We built the foundation for FABLE’s 800 question–answer pairs for travel routes by assembling 1,500 turn-by-turn travel routes, with three plans for each of 500 start–destination pairs. We began by compiling candidate endpoints from five datasets: the National Park Service’s listings of historic landmarks \cite{nps_national_register}, the San Diego Supercomputer Center’s registry of government buildings \cite{caloes_major_state_buildings}, the HIFLD “USA Hospitals” collection on Kaggle \cite{hifld_usa_hospitals}, the Bureau of Transportation Statistics’ “Airports—Citizen Connect” dataset \cite{bts_airports_citizen_connect}, and a community-curated dataset of airports in the US \cite{ourairports_usa}. From this pool, we randomly drew location pairs under two criteria: both sites lie within the same U.S. state (within the contiguous US) and the straight‑line distance between them does not exceed 150 km. If a sampled pair failed to produce three distinct routes, we discarded it and repeated sampling until 500 pairs were obtained.

For route generation, we tested several engines and ultimately selected Valhalla \cite{valhallaopen} for its granular natural-language maneuvers that include both action type and street name. Each start–destination pair was processed through Valhalla’s routing API using OpenStreetMap data \cite{OpenStreetMap} with five distinct cost configurations: default time-optimized, highway avoidance, severe maneuver penalty (1,000 s per maneuver), distance-only optimization, and residential-street preference. We then chose three routes with unique textual instructions. If fewer than three routes were unique, the start-destination pair was re-sampled to maintain uniformity.

Then, these results were checked for ambiguity and sequential coherence. We manually checked each maneuver type (e.g. Right, StayLeft, UturnLeft) under different conditions, such as turning onto unnamed or named streets and checked 2\% of the routes (10 start-destination pairs or 30 routes) to ensure that their natural language representations were coherent and could be followed sequentially to reach the destination. 

We then performed a two-stage normalization to eliminate instruction ambiguity and ensure sequential coherence. First, we flagged steps lacking clear spatial anchors or distance qualifiers (e.g., “Turn right.” or “Turn right onto Bees Ferry Road.” followed immediately by “You have arrived at your destination.”). These ambiguous steps were clarified by adding distances and durations in the natural language representations of the steps. Second, we resolved redundant roundabout instructions by merging the separate “Enter roundabout and take the 2nd exit onto Main Street” and “Exit the roundabout onto Main Street” into a single maneuver, consolidating the durations and road names from both steps.  The resulting steps were stored as both natural language and structured data.

To create the 800 QA pairs, we first defined a set of question templates based on the data present in the natural language descriptions of the steps. Using these templates, we identified sets of steps that matched the required patterns for each question type. We then generated ground-truth answers using the corresponding structured data, explicitly avoiding external knowledge, as all questions were required to have a definite answer grounded solely in the information provided within the route steps. Finally, we filtered this collection down to 100 QA pairs per domain, maximizing the number of unique routes present and the overall difficulty of each question. To ensure quality, 5\% of the QA pairs were manually validated for accuracy and clarity.

\subsection{Prompt Examples for Data-Flow Analyses across Domains}

Figures~\ref{fig:travel-reaching-definitions}--\ref{fig:travel-concurrency-analysis} present the prompt examples for all eight data-flow analyses in the travel routes domain. Figures~\ref{fig:plans-reaching-definitions}--\ref{fig:plans-concurrency-analysis} show the corresponding prompts for the automated plans domain. Finally, Figures~\ref{fig:recipes-reaching-definitions}--\ref{fig:recipes-concurrency-analysis} illustrate the prompts for the recipes domain. Each prompt encapsulates a unique benchmark analysis scenario and is designed to test the model's procedural reasoning capabilities under the constraints of a specific data-flow formalism.

\begin{figure}[htbp]
\centering
\scriptsize
\begin{tcolorbox}[
    colframe=black!0, 
    colback=gray!5, 
    coltitle=black,
    boxrule=0.5pt, 
    title=Travel Routes: Reaching Definitions, 
    fonttitle=\bfseries,
    width=\columnwidth, 
    rounded corners, 
    colbacktitle=gray!20
]
\textbf{You are an AI analyst being evaluated on data-flow analyses for travel routes.} \\
\textbf{Benchmark:} Reaching Definitions \\
\textbf{Definition:} Tracks which definitions (assignments) of variables can reach a particular program point without being overridden. \\

\textbf{Given the following travel planning scenario:} \\
\textbf{Goal:} Coit Memorial Tower (San Francisco, CA) to QUEEN OF THE VALLEY HOSPITAL - NAPA (NAPA, CA) \\
\textbf{Route steps:}
\begin{itemize}[leftmargin=1.5em]
  \item Drive north on Telegraph Hill Boulevard.
  \item Bear left onto Lombard Street.
  \item Turn left onto Columbus Avenue.
  \item Turn right onto Broadway.
  \item Turn right onto US 101 North/Van Ness Avenue.
  \item Turn left onto US 101 North/Lombard Street. Continue on US 101 North.
  \item Keep right to take CA 37 toward Napa.
  \item Take exit 19 on the right onto CA 29 toward Napa.
  \item Bear left onto CA 29/Sonoma Boulevard. Continue on CA 29.
  \item Keep left to stay on CA 29.
  \item Take exit 19 on the right toward Trancas Street/Redwood Road.
  \item Turn right onto Trancas Street.
  \item Turn left after 937 meters or 77--94 seconds.
  \item Turn right after 146 meters or 20--24 seconds.
  \item Continue for 21 meters or 8--9 seconds.
  \item Your destination is on the right.
\end{itemize}

\textbf{Question:} After Step 13, is the driver still on ‘CA 29‘ from Step 9? \\
\textbf{Instruction:} Please answer with only “Yes” or “No” without any further explanation. Adhere to this rule strictly.
\end{tcolorbox}
\caption{Prompt example for the Reaching Definitions analysis in the Travel Routes domain.}
\label{fig:travel-reaching-definitions}
\end{figure}

\begin{figure}[htbp]
\centering
\scriptsize
\begin{tcolorbox}[
    colframe=black!0, 
    colback=gray!5, 
    coltitle=black,
    boxrule=0.5pt, 
    title=Travel Routes: Very Busy Expressions, 
    fonttitle=\bfseries,
    width=\columnwidth, 
    rounded corners, 
    colbacktitle=gray!20
]
\textbf{You are an AI analyst being evaluated on data-flow analyses for travel routes.} \\
\textbf{Benchmark:} Very Busy Expressions \\
\textbf{Definition:} An expression is very busy if on every path forward from that point, the expression will be evaluated before any operand changes. \\

\textbf{Given the following travel planning scenario:} \\
\textbf{Goal:} McDonald, J. D., House (Fremont, NE) to NORFOLK REGIONAL CENTER (NORFOLK, NE) \\
\textbf{Route steps:}
\begin{itemize}[leftmargin=1.5em]
  \item Drive west on East Military Avenue.
  \item Turn right onto North Broad Street.
  \item Keep left to take Highway 275.
  \item Turn right onto Channel Road/NE 35/US 275 Business. Continue on NE 35/US 275 Business.
  \item Enter the roundabout and take the 2nd exit onto North Victory Road.
  \item Turn right after 1946 meters or 99--121 seconds.
  \item Turn right after 207 meters or 29--35 seconds.
  \item Turn left after 187 meters or 29--36 seconds.
  \item Turn left after 53 meters or 7--9 seconds.
  \item Continue for 48 meters or 8--9 seconds.
  \item You have arrived at your destination.
\end{itemize}

\textbf{Question:} Is Step 1 a very busy expression? \\
\textbf{Instruction:} Please answer with only “Yes” or “No” without any further explanation. Adhere to this rule strictly.
\end{tcolorbox}
\caption{Prompt example for the Very Busy Expressions analysis in the Travel Routes domain.}
\label{fig:travel-very-busy-expressions}
\end{figure}

\begin{figure}[htbp]
\centering
\scriptsize
\begin{tcolorbox}[
    colframe=black!0, 
    colback=gray!5, 
    coltitle=black,
    boxrule=0.5pt, 
    title=Travel Routes: Available Expressions, 
    fonttitle=\bfseries,
    width=\columnwidth, 
    rounded corners, 
    colbacktitle=gray!20
]
\textbf{You are an AI analyst being evaluated on data-flow analyses for travel routes.} \\
\textbf{Benchmark:} Available Expressions \\
\textbf{Definition:} An expression is available if it has already been computed and its operands have not been redefined since. \\

\textbf{Given the following travel planning scenario:} \\
\textbf{Goal:} VA MEDICAL CENTER - WEST ROXBURY DIVISION (WEST ROXBURY, MA) to Noyes, J.A., House (Cambridge, MA) \\
\textbf{Route steps:}
\begin{itemize}[leftmargin=1.5em]
  \item Drive south.
  \item Turn left after 117 meters or 19--23 seconds.
  \item Turn right onto Spring Street.
  \item Turn right onto Veterans of Foreign Wars Parkway.
  \item Bear left onto Baker Street.
  \item Bear left onto Dedham Street.
  \item Turn right onto Parker Street.
  \item Keep left to take Centre Street.
  \item Turn left onto Brattle Street.
  \item Turn left onto Fayerweather Street.
  \item Turn right onto Reservoir Street.
  \item Turn right onto Highland Street.
  \item Continue for 400 meters or 44--54 seconds.
  \item Your destination is on the left.
\end{itemize}

\textbf{Question:} Is Step 11 available in Step 12? \\
\textbf{Instruction:} Please answer with only “Yes” or “No” without any further explanation. Adhere to this rule strictly.
\end{tcolorbox}
\caption{Prompt example for the Available Expressions analysis in the Travel Routes domain.}
\label{fig:travel-available-expressions}
\end{figure}

\begin{figure}[htbp]
\centering
\scriptsize
\begin{tcolorbox}[
    colframe=black!0, 
    colback=gray!5, 
    coltitle=black,
    boxrule=0.5pt, 
    title=Travel Routes: Type-State Analysis, 
    fonttitle=\bfseries,
    width=\columnwidth, 
    rounded corners, 
    colbacktitle=gray!20
]
\textbf{You are an AI analyst being evaluated on data-flow analyses for travel routes.} \\
\textbf{Benchmark:} Type-State Analysis \\
\textbf{Definition:} Ensures an object is used only in valid states (e.g., a file must be opened before reading). \\

\textbf{Given the following travel planning scenario:} \\
\textbf{Goal:} Zanesville Municipal Airport (Zanesville, OH) to The Ohio State University Airport - Don Scott Field (Columbus, OH) \\
\textbf{Route steps:}
\begin{itemize}[leftmargin=1.5em]
  \item Drive northeast on Ritchey Parkway.
  \item Turn left onto Air Park Drive.
  \item Turn right onto Airport Road.
  \item Turn left to take the I 70 ramp.
  \item Keep right to take exit 108 onto I 270 North.
  \item Keep right to take I 270 North.
  \item Keep left to take I 270 North toward Main Street/Whitehall.
  \item Keep left to take I 270 North/Jack Nicklaus Freeway.
  \item Keep left to stay on Jack Nicklaus Freeway.
  \item Take exit 20 on the right toward Sawmill Road.
  \item Keep left to take Sawmill Road/CR 70 toward Sawmill Road South.
  \item Turn left after 3426 meters or 178--217 seconds.
  \item Continue for 1238 meters or 400--489 seconds.
  \item Your destination is on the left.
\end{itemize}

\textbf{Question:} Can Step 5 be performed without completing Step 4? \\
\textbf{Instruction:} Please answer with only “Yes” or “No” without any further explanation. Adhere to this rule strictly.
\end{tcolorbox}
\caption{Prompt example for the Type-State Analysis in the Travel Routes domain.}
\label{fig:travel-type-state-analysis}
\end{figure}

\begin{figure}[htbp]
\centering
\scriptsize
\begin{tcolorbox}[
    colframe=black!0, 
    colback=gray!5, 
    coltitle=black,
    boxrule=0.5pt, 
    title=Travel Routes: Live Variable Analysis, 
    fonttitle=\bfseries,
    width=\columnwidth, 
    rounded corners, 
    colbacktitle=gray!20
]
\textbf{You are an AI analyst being evaluated on data-flow analyses for travel routes.} \\
\textbf{Benchmark:} Live Variable Analysis \\
\textbf{Definition:} A variable is live if its value will be used again in the future (i.e., before being overwritten or discarded). \\

\textbf{Given the following travel planning scenario:} \\
\textbf{Goal:} Hartford Club (Hartford, CT) to Igor I Sikorsky Memorial Airport (Bridgeport, CT) \\
\textbf{Route steps:}
\begin{itemize}[leftmargin=1.5em]
  \item Drive east on Prospect Street.
  \item Turn right onto US 5/Main Street.
  \item Turn right onto US 5/East River Drive.
  \item Turn left to take the US 5 ramp.
  \item Take exit 86 on the right onto I 91 South toward New Haven/New York City.
  \item Take exit 17 on the right onto CT 15 South/Wilbur Cross Parkway.
  \item Keep left to take CT 15 South/Wilbur Cross Parkway.
  \item Take exit 37 on the right toward I 95/US 1/Milford.
  \item Take exit 2B on the right onto I 95 South toward Bridgeport.
  \item Take exit 30 on the right onto CT 113 toward Surf Avenue.
  \item Turn left onto Surf Avenue.
  \item Turn left onto Lordship Boulevard/CT 113.
  \item Turn right to stay on Lordship Boulevard/CT 113.
  \item Turn left onto Great Meadow Road.
  \item Continue for 622 meters or 59--73 seconds.
  \item Your destination is on the right.
\end{itemize}

\textbf{Question:} Is the effect from Step 2, the driver getting on US 5, live at any point after Step 2? Please answer with only “Yes” or “No” without any further explanation. Adhere to this rule strictly.
\end{tcolorbox}
\caption{Prompt example for the Live Variable Analysis in the Travel Routes domain.}
\label{fig:travel-live-variable-analysis}
\end{figure}

\begin{figure}[htbp]
\centering
\scriptsize
\begin{tcolorbox}[
    colframe=black!0, 
    colback=gray!5, 
    coltitle=black,
    boxrule=0.5pt, 
    title=Travel Routes: Interval Analysis, 
    fonttitle=\bfseries,
    width=\columnwidth, 
    rounded corners, 
    colbacktitle=gray!20
]
\textbf{You are an AI analyst being evaluated on data-flow analyses for travel routes.} \\
\textbf{Benchmark:} Interval Analysis \\
\textbf{Definition:} Determines the possible numeric ranges (e.g., bounds on variables) at each point in a program. \\

\textbf{Given the following travel planning scenario:} \\
\textbf{Goal:} HENRY FORD WEST BLOOMFIELD HOSPITAL (W Bloomfield, MI) to Willow Run Airport (Detroit, MI) \\
\textbf{Route steps:}
\begin{itemize}[leftmargin=1.5em]
  \item Drive east.
  \item Turn right after 221 meters or 29--35 seconds.
  \item Turn left after 24 meters or 8--10 seconds.
  \item Turn left onto West Maple Road.
  \item Turn left onto Haggerty Road.
  \item Turn left onto 8 Mile Road.
  \item Take the I 275 South ramp on the right toward Detroit/Toledo.
  \item Take exit 20 on the right onto Ecorse Road toward Romulus/Willow Run Airport.
  \item Turn right onto Ecorse Road.
  \item Turn left after 7078 meters or 283--346 seconds.
  \item Turn left after 78 meters or 39--48 seconds.
  \item Turn right after 3165 meters or 412--504 seconds.
  \item Continue for 806 meters or 109--134 seconds.
  \item Your destination is on the right.
\end{itemize}

\textbf{Question:} What is the time interval between performing Step 9 and Step 13? \\
\textbf{Instruction:} Please provide the numeric interval in the format [x, y] without any further explanation.
\end{tcolorbox}
\caption{Prompt example for the Interval Analysis in the Travel Routes domain.}
\label{fig:travel-interval-analysis}
\end{figure}

\begin{figure}[htbp]
\centering
\scriptsize
\begin{tcolorbox}[
    colframe=black!0, 
    colback=gray!5, 
    coltitle=black,
    boxrule=0.5pt, 
    title=Travel Routes: Taint Analysis, 
    fonttitle=\bfseries,
    width=\columnwidth, 
    rounded corners, 
    colbacktitle=gray!20
]
\textbf{You are an AI analyst being evaluated on data-flow analyses for travel routes.} \\
\textbf{Benchmark:} Taint Analysis \\
\textbf{Definition:} Tracks tainted or untrusted data to ensure it does not reach sensitive areas without sanitization. \\

\textbf{Given the following travel planning scenario:} \\
\textbf{Goal:} SAINT FRANCIS HOSPITAL (Roslyn, NY) to NYACK HOSPITAL (Nyack, NY) \\
\textbf{Route steps:}
\begin{itemize}[leftmargin=1.5em]
  \item Drive north.
  \item Turn right after 13 meters or 3 seconds.
  \item Turn right onto Port Washington Boulevard/NY 101.
  \item Turn right onto North Service Road.
  \item Take the I 495 West ramp toward New York.
  \item Take exit 31N-S on the right toward Whitestone Bridge.
  \item Keep right to take exit 31N onto CI/Cross Island Parkway/100th Infantry Division Parkway toward Whitestone Bridge.
  \item Keep right to take exit 33 onto I 295 North toward Throgs Neck Bridge/Bronx/New England.
  \item Keep left to take Cross Bronx Expressway.
  \item Keep left toward Upper Level.
  \item Take exit 74 on the right toward Palisades Parkway.
  \item Take exit 4 on the right onto US 9W.
  \item Turn left onto Palisades Boulevard/US 9W. Continue on US 9W.
  \item Turn left to stay on US 9W.
  \item Turn right onto Hillside Avenue/US 9W. Continue on US 9W.
  \item Turn right after 1623 meters or 79--97 seconds.
  \item Continue for 48 meters or 11--14 seconds.
  \item You have arrived at your destination.
\end{itemize}

\textbf{Question:} Does turning left in Step 16 taint the preconditions for the rest of the directions? \\
\textbf{Instruction:} Please answer with only “Yes” or “No” without any further explanation. Adhere to this rule strictly.
\end{tcolorbox}
\caption{Prompt example for the Taint Analysis in the Travel Routes domain.}
\label{fig:travel-taint-analysis}
\end{figure}

\begin{figure}[htbp]
\centering
\scriptsize
\begin{tcolorbox}[
    colframe=black!0, 
    colback=gray!5, 
    coltitle=black,
    boxrule=0.5pt, 
    title=Travel Routes: Concurrency Analysis, 
    fonttitle=\bfseries,
    width=\columnwidth, 
    rounded corners, 
    colbacktitle=gray!20
]
\textbf{You are an AI analyst being evaluated on data-flow analyses for travel routes.} \\
\textbf{Benchmark:} Concurrency Analysis \\
\textbf{Definition:} Examines parallel (concurrent) execution paths to detect data races or synchronization issues. \\

\textbf{Given the following travel planning scenario:} \\
\textbf{Goal:} House at 220 Walnut Street (Nogales, AZ) to BENSON HOSPITAL (Benson, AZ) \\
\textbf{Route steps:}
\begin{itemize}[leftmargin=1.5em]
  \item Drive east on West Oak Street.
  \item Turn left onto North Grand Avenue/I 19 Bus.
  \item Make a sharp left to stay on North Grand Avenue/I 19 Bus.
  \item Turn right onto AZ 82.
  \item Turn left onto AZ 90/State Route 90. Continue on AZ 90.
  \item Turn right to take the ramp toward I 10/Benson/Douglas.
  \item Take exit 303 on the right onto I 10 Business toward AZ 80 East/Tombstone/Douglas.
  \item Turn right onto South Ocotillo Avenue.
  \item Turn right after 569 meters or 42--52 seconds.
  \item Continue for 71 meters or 14--18 seconds.
  \item Your destination is on the right.
\end{itemize}

\textbf{Question:} Can Steps 10 and 6 be performed concurrently? \\
\textbf{Instruction:} Please answer with only “Yes” or “No” without any further explanation. Adhere to this rule strictly.
\end{tcolorbox}
\caption{Prompt example for the Concurrency Analysis in the Travel Routes domain.}
\label{fig:travel-concurrency-analysis}
\end{figure}

\begin{figure}[h!]
\centering
\begin{tcolorbox}[
    colframe=black!0, 
    colback=teal!5, 
    coltitle=black,
    boxrule=0.5pt, 
    title=Plans: Reaching Definitions, 
    fonttitle=\bfseries,
    width=\columnwidth, 
    rounded corners, 
    colbacktitle=teal!20
]
\textbf{You are an AI analyst being evaluated on data-flow analyses for planning tasks.} \\
\textbf{Benchmark:} Reaching Definitions \\
\textbf{Definition:} Tracks which definitions (assignments) of variables can reach a particular program point without being overridden. \\

\textbf{Given the following planning task:} \\
\textbf{Goal:} \texttt{(and (at c0 l3) (at c1 l1) (at c2 l2))} \\
\textbf{Plan steps:}
\begin{itemize}[leftmargin=1.5em]
  \item[1:] \texttt{board(c2, l1)}
  \item[2:] \texttt{sail(l1, l2)}
  \item[3:] \texttt{debark(c2, l2)}
  \item[4:] \texttt{sail(l2, l3)}
  \item[5:] \texttt{board(c1, l3)}
  \item[6:] \texttt{sail(l3, l1)}
  \item[7:] \texttt{debark(c1, l1)}
\end{itemize}

\textbf{Question:} In Step 3 (\texttt{debark}), is the predicate \texttt{('on', 'c2')} potentially from the effect of Step 1 (\texttt{board}) being used? \\
\textbf{Instruction:} Please answer with only “Yes” or “No” without any further explanation. Adhere to this rule strictly.
\end{tcolorbox}
\caption{Prompt example for the Reaching Definitions analysis in the Plans domain.}
\label{fig:plans-reaching-definitions}
\end{figure}

\begin{figure}[h!]
\centering
\begin{tcolorbox}[
    colframe=black!0, 
    colback=teal!5, 
    coltitle=black,
    boxrule=0.5pt, 
    title=Plans: Very Busy Expressions, 
    fonttitle=\bfseries,
    width=\columnwidth, 
    rounded corners, 
    colbacktitle=teal!20
]
\textbf{You are an AI analyst being evaluated on data-flow analyses for planning tasks.} \\
\textbf{Benchmark:} Very Busy Expressions \\
\textbf{Definition:} An expression is very busy if on every path forward from that point, the expression will be evaluated before any operand changes. \\

\textbf{Given the following planning task:} \\
\textbf{Goal:} \texttt{(and (at c0 l1) (at c1 l3) (at c2 l2))} \\
\textbf{Plan steps:}
\begin{itemize}[leftmargin=1.5em]
  \item[1:] \texttt{board(c2, l1)}
  \item[2:] \texttt{sail(l1, l2)}
  \item[3:] \texttt{debark(c2, l2)}
  \item[4:] \texttt{board(c1, l2)}
  \item[5:] \texttt{sail(l2, l3)}
  \item[6:] \texttt{debark(c1, l3)}
\end{itemize}

\textbf{Question:} Is the action in Step 2 (\texttt{sail(l1, l2)}) \emph{very busy} in the sense that its effect \texttt{('at-ferry', 'l2')} is used by the next step, Step 3 (\texttt{debark(c2, l2)})? \\
\textbf{Instruction:} Please answer with only “Yes” or “No” without any further explanation. Adhere to this rule strictly.
\end{tcolorbox}
\caption{Prompt example for the Very Busy Expressions analysis in the Plans domain.}
\label{fig:plans-very-busy-expressions}
\end{figure}

\begin{figure}[h!]
\centering
\begin{tcolorbox}[
    colframe=black!0, 
    colback=teal!5, 
    coltitle=black,
    boxrule=0.5pt, 
    title=Plans: Available Expressions, 
    fonttitle=\bfseries,
    width=\columnwidth, 
    rounded corners, 
    colbacktitle=teal!20
]
\textbf{You are an AI analyst being evaluated on data-flow analyses for planning tasks.} \\
\textbf{Benchmark:} Available Expressions \\
\textbf{Definition:} An expression is available if it has already been computed and its operands have not been redefined since. \\

\textbf{Given the following planning task:} \\
\textbf{Goal:} \texttt{(and (at package1 s1) (at package2 s3))} \\
\textbf{Plan steps:}
\begin{itemize}[leftmargin=1.5em]
  \item[1:] \texttt{board-truck(driver2, truck1, s1)}
  \item[2:] \texttt{load-truck(package2, truck1, s1)}
  \item[3:] \texttt{drive-truck(truck1, s1, s2, driver2)}
  \item[4:] \texttt{load-truck(package1, truck1, s2)}
  \item[5:] \texttt{drive-truck(truck1, s2, s1, driver2)}
  \item[6:] \texttt{unload-truck(package1, truck1, s1)}
  \item[7:] \texttt{drive-truck(truck1, s1, s3, driver2)}
  \item[8:] \texttt{unload-truck(package2, truck1, s3)}
\end{itemize}

\textbf{Question:} Is the effect \texttt{('driving', 'driver2', 'truck1')} from Step 1 (\texttt{board-truck}) still available for Step 3 (\texttt{drive-truck})? \\
\textbf{Instruction:} Please answer with only “Yes” or “No” without any further explanation. Adhere to this rule strictly.
\end{tcolorbox}
\caption{Prompt example for the Available Expressions analysis in the Plans domain.}
\label{fig:plans-available-expressions}
\end{figure}

\begin{figure}[h!]
\centering
\begin{tcolorbox}[
    colframe=black!0, 
    colback=teal!5, 
    coltitle=black,
    boxrule=0.5pt, 
    title=Plans: Live Variable Analysis, 
    fonttitle=\bfseries,
    width=\columnwidth, 
    rounded corners, 
    colbacktitle=teal!20
]
\textbf{You are an AI analyst being evaluated on data-flow analyses for planning tasks.} \\
\textbf{Benchmark:} Live Variable Analysis \\
\textbf{Definition:} A variable is live if its value will be used again in the future (i.e., before being overwritten or discarded). \\

\textbf{Given the following planning task:} \\
\textbf{Goal:} \texttt{(and (at package1 s1) (at package2 s3))} \\
\textbf{Plan steps:}
\begin{itemize}[leftmargin=1.5em]
  \item[1:] \texttt{board-truck(driver2, truck1, s1)}
  \item[2:] \texttt{load-truck(package2, truck1, s1)}
  \item[3:] \texttt{drive-truck(truck1, s1, s2, driver2)}
  \item[4:] \texttt{load-truck(package1, truck1, s2)}
  \item[5:] \texttt{drive-truck(truck1, s2, s1, driver2)}
  \item[6:] \texttt{unload-truck(package1, truck1, s1)}
  \item[7:] \texttt{drive-truck(truck1, s1, s3, driver2)}
  \item[8:] \texttt{unload-truck(package2, truck1, s3)}
\end{itemize}

\textbf{Question:} After Step 1 (\texttt{board-truck}), is the effect \texttt{('driving', 'driver2', 'truck1')} \emph{live} (i.e., needed by a future step like Step 3)? \\
\textbf{Instruction:} Please answer with only “Yes” or “No” without any further explanation. Adhere to this rule strictly.
\end{tcolorbox}
\caption{Prompt example for the Live Variable Analysis in the Plans domain.}
\label{fig:plans-live-variable-analysis}
\end{figure}

\begin{figure}[h!]
\centering
\begin{tcolorbox}[
    colframe=black!0, 
    colback=teal!5, 
    coltitle=black,
    boxrule=0.5pt, 
    title=Plans: Interval Analysis, 
    fonttitle=\bfseries,
    width=\columnwidth, 
    rounded corners, 
    colbacktitle=teal!20
]
\textbf{You are an AI analyst being evaluated on data-flow analyses for planning tasks.} \\
\textbf{Benchmark:} Interval Analysis \\
\textbf{Definition:} Determines the possible numeric ranges (e.g., bounds on variables) at each point in a program. \\

\textbf{Given the following planning task:} \\
\textbf{Goal:} \texttt{(and (at c0 l1) (at c1 l3) (at c2 l2))} \\
\textbf{Plan steps:}
\begin{itemize}[leftmargin=1.5em]
  \item[1:] \texttt{board(c2, l1)}
  \item[2:] \texttt{sail(l1, l2)}
  \item[3:] \texttt{debark(c2, l2)}
  \item[4:] \texttt{board(c1, l2)}
  \item[5:] \texttt{sail(l2, l3)}
  \item[6:] \texttt{debark(c1, l3)}
\end{itemize}

\textbf{Question:} Based on immediate dependencies in this linear plan, what constraints determine when Step 6 (\texttt{debark(c1, l3)}) must occur? \\
\textbf{Instruction:} Please answer \emph{before}, \emph{after}, or \emph{between which steps} without any further explanation.
\end{tcolorbox}
\caption{Prompt example for the Interval Analysis in the Plans domain.}
\label{fig:plans-interval-analysis}
\end{figure}

\begin{figure}[h!]
\centering
\begin{tcolorbox}[
    colframe=black!0, 
    colback=teal!5, 
    coltitle=black,
    boxrule=0.5pt, 
    title=Plans: Type-State Analysis, 
    fonttitle=\bfseries,
    width=\columnwidth, 
    rounded corners, 
    colbacktitle=teal!20
]
\textbf{You are an AI analyst being evaluated on data-flow analyses for planning tasks.} \\
\textbf{Benchmark:} Type-State Analysis \\
\textbf{Definition:} Ensures an object is used only in valid states (e.g., a file must be opened before reading). \\

\textbf{Given the following planning task:} \\
\textbf{Goal:} \texttt{(and (on d1 d2) (clear d1) (on d2 peg1) (clear peg2) (clear peg3))} \\
\textbf{Plan steps:}
\begin{itemize}[leftmargin=1.5em]
  \item[1:] \texttt{move(d1, d2, peg3)}
  \item[2:] \texttt{move(d2, peg2, peg1)}
  \item[3:] \texttt{move(d1, peg3, d2)}
\end{itemize}

\textbf{Question:} If Step 1 (\texttt{move}) were skipped, would Step 2 (\texttt{move}) become invalid? \\
\textbf{Instruction:} Please answer with only “Yes” or “No” without any further explanation. Adhere to this rule strictly.
\end{tcolorbox}
\caption{Prompt example for the Type-State Analysis in the Plans domain.}
\label{fig:plans-type-state-analysis}
\end{figure}

\begin{figure}[h!]
\centering
\begin{tcolorbox}[
    colframe=black!0, 
    colback=teal!5, 
    coltitle=black,
    boxrule=0.5pt, 
    title=Plans: Taint Analysis, 
    fonttitle=\bfseries,
    width=\columnwidth, 
    rounded corners, 
    colbacktitle=teal!20
]
\textbf{You are an AI analyst being evaluated on data-flow analyses for planning tasks.} \\
\textbf{Benchmark:} Taint Analysis \\
\textbf{Definition:} Tracks tainted or untrusted data to ensure it does not reach sensitive areas without sanitization. \\

\textbf{Given the following planning task:} \\
\textbf{Goal:} \texttt{(and (at c0 l1) (at c1 l3) (at c2 l2))} \\
\textbf{Plan steps:}
\begin{itemize}[leftmargin=1.5em]
  \item[1:] \texttt{board(c2, l1)}
  \item[2:] \texttt{sail(l1, l2)}
  \item[3:] \texttt{debark(c2, l2)}
  \item[4:] \texttt{board(c1, l2)}
  \item[5:] \texttt{sail(l2, l3)}
  \item[6:] \texttt{debark(c1, l3)}
\end{itemize}

\textbf{Question:} Does any step in this plan interfere with (taint) the preconditions of the immediately following step? \\
\textbf{Instruction:} Please answer with only “Yes” or “No” without any further explanation. Adhere to this rule strictly.
\end{tcolorbox}
\caption{Prompt example for the Taint Analysis in the Plans domain.}
\label{fig:plans-taint-analysis}
\end{figure}

\begin{figure}[h!]
\centering
\begin{tcolorbox}[
    colframe=black!0, 
    colback=teal!5, 
    coltitle=black,
    boxrule=0.5pt, 
    title=Plans: Concurrency Analysis, 
    fonttitle=\bfseries,
    width=\columnwidth, 
    rounded corners, 
    colbacktitle=teal!20
]
\textbf{You are an AI analyst being evaluated on data-flow analyses for planning tasks.} \\
\textbf{Benchmark:} Concurrency Analysis \\
\textbf{Definition:} Examines parallel (concurrent) execution paths to detect data races or synchronization issues. \\

\textbf{Given the following planning task:} \\
\textbf{Goal:} \texttt{(and (at c0 l1) (at c1 l3) (at c2 l2))} \\
\textbf{Plan steps:}
\begin{itemize}[leftmargin=1.5em]
  \item[1:] \texttt{board(c2, l1)}
  \item[2:] \texttt{sail(l1, l2)}
  \item[3:] \texttt{debark(c2, l2)}
  \item[4:] \texttt{board(c1, l2)}
  \item[5:] \texttt{sail(l2, l3)}
  \item[6:] \texttt{debark(c1, l3)}
\end{itemize}

\textbf{Question:} Based on simple precondition/delete analysis, can Step 4 (\texttt{board}) and Step 3 (\texttt{debark}) run concurrently? \\
\textbf{Instruction:} Please answer with only “Yes” or “No” without any further explanation. Adhere to this rule strictly.
\end{tcolorbox}
\caption{Prompt example for the Concurrency Analysis in the Plans domain.}
\label{fig:plans-concurrency-analysis}
\end{figure}

\begin{figure}[h!]
\centering
\begin{tcolorbox}[
    colframe=black!0, 
    colback=orange!5, 
    coltitle=black,
    boxrule=0.5pt, 
    title=Recipes: Reaching Definitions, 
    fonttitle=\bfseries,
    width=\columnwidth, 
    rounded corners, 
    colbacktitle=orange!20
]
\textbf{You are an AI analyst being evaluated on data-flow analyses for cooking recipes.} \\
\textbf{Benchmark:} Reaching Definitions \\
\textbf{Definition:} Tracks which definitions (assignments) of variables can reach a particular program point without being overridden. \\

\textbf{Given the following recipe scenario:} \\
\textbf{Goal:} Amish Friendship Cake \\
\textbf{Steps:}
\begin{itemize}[leftmargin=1.5em]
  \item[1:] Preheat the oven to 350 degrees F (175 degrees C). Grease two 9-inch loaf pans.
  \item[2:] Mix flour, sourdough starter, sugar, oil, eggs, baking powder, vanilla, and salt together in a bowl. Stir in chopped walnuts. Stir in baking soda and pour batter into the prepared pans.
  \item[3:] Bake in the preheated oven until a toothpick inserted into the center comes out clean, 45 minutes to 1 hour.
\end{itemize}

\textbf{Question:} In Step 3, is the oven from Step 1 being used? \\
\textbf{Instruction:} Please answer with only “Yes” or “No” without any further explanation. Adhere to this rule strictly.
\end{tcolorbox}
\caption{Prompt example for the Reaching Definitions analysis in the Recipes domain.}
\label{fig:recipes-reaching-definitions}
\end{figure}

\begin{figure}[h!]
\centering
\begin{tcolorbox}[
    colframe=black!0, 
    colback=orange!5, 
    coltitle=black,
    boxrule=0.5pt, 
    title=Recipes: Very Busy Expressions, 
    fonttitle=\bfseries,
    width=\columnwidth, 
    rounded corners, 
    colbacktitle=orange!20
]
\textbf{You are an AI analyst being evaluated on data-flow analyses for cooking recipes.} \\
\textbf{Benchmark:} Very Busy Expressions \\
\textbf{Definition:} An expression is very busy if on every path forward from that point, the expression will be evaluated before any operand changes. \\

\textbf{Given the following recipe scenario:} \\
\textbf{Goal:} Bulgur Wheat Pilaf \\
\textbf{Steps:}
\begin{itemize}[leftmargin=1.5em]
  \item[1:] Place the eggs in a saucepan of cold water, bring to a boil, and boil gently for 10 minutes. Drain the eggs and cool under cold running water. Set aside.
  \item[2:] While the eggs are cooking, heat the oil in a large saucepan over medium-high heat. Add the onion and garlic, and saute for 3 minutes, stirring occasionally. Stir in the ground coriander, cinnamon, turmeric, and chilies (if using). Stir for 1 minute.
  \item[3:] Add the pinto beans or chickpeas and apricots, and stir to coat them with the spices. Stir in the bulgur and green beans, then pour in enough water to cover by about 1/2 inch. Bring to a boil, then reduce the heat to low. Cover and simmer for 20 minutes, or until all of the liquid has been absorbed.
  \item[4:] While the bulgur is cooking, shell and slice the eggs. Fluff the bulgur with a fork and season lightly with salt and pepper. Serve hot, garnished with the egg slices and sprinkled with cilantro leaves.
\end{itemize}

\textbf{Question:} Is \texttt{saucepan} from Step 1 used in multiple future steps without being redefined? \\
\textbf{Instruction:} Please answer with only “Yes” or “No” without any further explanation. Adhere to this rule strictly.
\end{tcolorbox}
\caption{Prompt example for the Very Busy Expressions analysis in the Recipes domain.}
\label{fig:recipes-very-busy-expressions}
\end{figure}

\begin{figure}[h!]
\centering
\begin{tcolorbox}[
    colframe=black!0, 
    colback=orange!5, 
    coltitle=black,
    boxrule=0.5pt, 
    title=Recipes: Available Expressions, 
    fonttitle=\bfseries,
    width=\columnwidth, 
    rounded corners, 
    colbacktitle=orange!20
]
\textbf{You are an AI analyst being evaluated on data-flow analyses for cooking recipes.} \\
\textbf{Benchmark:} Available Expressions \\
\textbf{Definition:} An expression is available if it has already been computed and its operands have not been redefined since. \\

\textbf{Given the following recipe scenario:} \\
\textbf{Goal:} Dark Chocolate Cookies (Gluten Free, Vegan) \\
\textbf{Steps:}
\begin{itemize}[leftmargin=1.5em]
  \item[1:] Preheat oven to 350 degrees F (175 degrees C). Line baking sheets with parchment paper.
  \item[2:] Combine 1 1/2 cups hot water, chia seeds, and flax seeds in a blender; let soak, about 10 minutes. Blend mixture on high until combined, about 1 minute.
  \item[3:] Combine millet flour, raw sugar, cacao powder, tofu, 1/2 cup warm water, agave syrup, 1/2 cup cacao nibs, almond meal, vanilla extract, baking soda, and salt in a large bowl. Fold in raspberries. Pour blended mixture over flour mixture; mix until batter is combined.
  \item[4:] Scoop batter out onto baking sheets. Top each scoop with a few cacao nibs.
  \item[5:] Bake in the preheated oven until firm, about 12 minutes. Transfer to wire racks to cool.
\end{itemize}

\textbf{Question:} Is \texttt{oven} from Step 1 still available in Step 2? \\
\textbf{Instruction:} Please answer with only “Yes” or “No” without any further explanation. Adhere to this rule strictly.
\end{tcolorbox}
\caption{Prompt example for the Available Expressions analysis in the Recipes domain.}
\label{fig:recipes-available-expressions}
\end{figure}

\begin{figure}[h!]
\centering
\begin{tcolorbox}[
    colframe=black!0, 
    colback=orange!5, 
    coltitle=black,
    boxrule=0.5pt, 
    title=Recipes: Live Variable Analysis, 
    fonttitle=\bfseries,
    width=\columnwidth, 
    rounded corners, 
    colbacktitle=orange!20
]
\textbf{You are an AI analyst being evaluated on data-flow analyses for cooking recipes.} \\
\textbf{Benchmark:} Live Variable Analysis \\
\textbf{Definition:} A variable is live if its value will be used again in the future (i.e., before being overwritten or discarded). \\

\textbf{Given the following recipe scenario:} \\
\textbf{Goal:} Crab and Cheddar Quiche \\
\textbf{Steps:}
\begin{itemize}[leftmargin=1.5em]
  \item[1:] Preheat the oven to 350 degrees F (175 degrees C).
  \item[2:] Beat eggs, mayonnaise, milk, flour, and 1 teaspoon Old Bay seasoning together in a bowl until smooth and creamy; stir in Cheddar cheese and parsley. Gently fold in crabmeat.
  \item[3:] Pour crabmeat filling into pie crust; sprinkle 1 pinch Old Bay seasoning over quiche.
  \item[4:] Bake in the preheated oven until a knife inserted into center comes out clean and top is lightly browned, 40 to 45 minutes.
\end{itemize}

\textbf{Question:} Is \texttt{eggs} live after Step 3? \\
\textbf{Instruction:} Please answer with only “Yes” or “No” without any further explanation. Adhere to this rule strictly.
\end{tcolorbox}
\caption{Prompt example for the Live Variable Analysis in the Recipes domain.}
\label{fig:recipes-live-variable-analysis}
\end{figure}

\begin{figure}[h!]
\centering
\scriptsize
\begin{tcolorbox}[
    colframe=black!0, 
    colback=orange!5, 
    coltitle=black,
    boxrule=0.5pt, 
    title=Recipes: Interval Analysis, 
    fonttitle=\bfseries,
    width=\columnwidth, 
    rounded corners, 
    colbacktitle=orange!20
]
\textbf{You are an AI analyst being evaluated on data-flow analyses for cooking recipes.} \\
\textbf{Benchmark:} Interval Analysis \\
\textbf{Definition:} Determines the possible numeric ranges (e.g., bounds on variables) at each point in a program. \\

\textbf{Given the following recipe scenario:} \\
\textbf{Goal:} Cornmeal Cinnamon Rolls \\
\textbf{Steps:}
\begin{itemize}[leftmargin=1.5em]
  \item[1:] In a saucepan, combine milk and cornmeal. Bring to a boil over medium heat, stirring constantly. Add the sugar, butter and salt; cool to 110 degrees F--115 degrees F. In a mixing bowl, dissolve yeast in warm water. Add the eggs, cornmeal mixture and 2 cups flour; beat until smooth. Stir in enough remaining flour to form a soft dough. Turn onto a floured surface; knead until smooth and elastic, about 6--8 minutes. Place in a greased bowl, turning once to grease top. Cover and let rise in a warm place until doubled, about 1 hour.
  \item[2:] Meanwhile, place raisins in a saucepan and cover with water. Bring to a boil; remove from heat. Cover and let stand for 15 minutes. Drain; set aside. Punch dough down. Turn onto a lightly floured surface; divide in half. Roll each into a 12-in. x 10-in. rectangle; brush with melted butter. Combine sugar, cinnamon and nutmeg if desired; sprinkle over dough to within 1/2 in. of edges. Sprinkle with raisins. Roll up, jelly-roll style, starting with a long side; pinch seam to seal. Cut each into 12 slices. Place, cut side down, in two greased 13-in. x 9-in. x 2-in. baking pans. Cover and let rise until doubled, about 1 hour. Bake at 375 degrees F for 20--25 minutes or until golden brown. Cool in pans on wire racks.
  \item[3:] For frosting, combine sugar, butter, cream cheese, extract and enough milk to achieve desired consistency. Spread or drizzle over rolls. Refrigerate leftovers.
\end{itemize}

\textbf{Question:} What is the last time interval specified in Step 1? \\
\textbf{Instruction:} Please provide the numeric interval exactly as it appears (e.g., ``1/2 hour'') without any further explanation.
\end{tcolorbox}
\caption{Prompt example for the Interval Analysis in the Recipes domain.}
\label{fig:recipes-interval-analysis}
\end{figure}

\begin{figure}[h!]
\centering
\begin{tcolorbox}[
    colframe=black!0, 
    colback=orange!5, 
    coltitle=black,
    boxrule=0.5pt, 
    title=Recipes: Type-State Analysis, 
    fonttitle=\bfseries,
    width=\columnwidth, 
    rounded corners, 
    colbacktitle=orange!20
]
\textbf{You are an AI analyst being evaluated on data-flow analyses for cooking recipes.} \\
\textbf{Benchmark:} Type-State Analysis \\
\textbf{Definition:} Ensures an object is used only in valid states (e.g., a file must be opened before reading). \\

\textbf{Given the following recipe scenario:} \\
\textbf{Goal:} Brazil Nut Fruitcake \\
\textbf{Steps:}
\begin{itemize}[leftmargin=1.5em]
  \item[1:] Preheat oven to 350 degrees F (175 degrees C). Grease 3 - 8x4 inch loaf pans and line them with parchment or waxed paper.
  \item[2:] Beat eggs, salt and vanilla together until very light and lemon colored. Stir in sugar, 1 cup flour and baking powder.
  \item[3:] Place cherries, nuts, and dates into a large bowl. Dust with the remaining 1/2 cup flour. Then stir in sugar mixture. There is very little batter which makes this a very stiff mixture. Mix with hands.
  \item[4:] Press batter into prepared loaf pans. Bake for 1 hour.
\end{itemize}

\textbf{Question:} If we skip Step 2, is it still valid to use the \texttt{sugar} in Step 3? \\
\textbf{Instruction:} Please answer with only “Yes” or “No” without any further explanation. Adhere to this rule strictly.
\end{tcolorbox}
\caption{Prompt example for the Type-State Analysis in the Recipes domain.}
\label{fig:recipes-type-state-analysis}
\end{figure}

\begin{figure}[h!]
\centering
\begin{tcolorbox}[
    colframe=black!0, 
    colback=orange!5, 
    coltitle=black,
    boxrule=0.5pt, 
    title=Recipes: Taint Analysis, 
    fonttitle=\bfseries,
    width=\columnwidth, 
    rounded corners, 
    colbacktitle=orange!20
]
\textbf{You are an AI analyst being evaluated on data-flow analyses for cooking recipes.} \\
\textbf{Benchmark:} Taint Analysis \\
\textbf{Definition:} Tracks tainted or untrusted data to ensure it does not reach sensitive areas without sanitization. \\

\textbf{Given the following recipe scenario:} \\
\textbf{Goal:} Carrot Cake \\
\textbf{Steps:}
\begin{itemize}[leftmargin=1.5em]
  \item[1:] Preheat oven to 350 degrees F (175 degrees C). Grease and flour a 9x13 inch pan. Sift together the flour, baking soda, salt and cinnamon. Set aside.
  \item[2:] In a large bowl, beat eggs, sugar and oil until smooth. Beat in flour mixture. Stir in pureed carrots, pineapple, walnuts and coconut. Pour batter into prepared pan.
  \item[3:] Bake in the preheated oven for 30 to 40 minutes, or until a toothpick inserted into the center of the cake comes out clean. Allow to cool.
\end{itemize}

\textbf{Question:} Does using \texttt{eggs} in Step 2 introduce a potential safety concern to the recipe? \\
\textbf{Instruction:} Please answer with only “Yes” or “No” without any further explanation. Adhere to this rule strictly.
\end{tcolorbox}
\caption{Prompt example for the Taint Analysis in the Recipes domain.}
\label{fig:recipes-taint-analysis}
\end{figure}

\begin{figure}[h!]
\centering
\begin{tcolorbox}[
    colframe=black!0, 
    colback=orange!5, 
    coltitle=black,
    boxrule=0.5pt, 
    title=Recipes: Concurrency Analysis, 
    fonttitle=\bfseries,
    width=\columnwidth, 
    rounded corners, 
    colbacktitle=orange!20
]
\textbf{You are an AI analyst being evaluated on data-flow analyses for cooking recipes.} \\
\textbf{Benchmark:} Concurrency Analysis \\
\textbf{Definition:} Examines parallel (concurrent) execution paths to detect data races or synchronization issues. \\

\textbf{Given the following recipe scenario:} \\
\textbf{Goal:} Classic Pizza Margherita \\
\textbf{Steps:}
\begin{itemize}[leftmargin=1.5em]
  \item[1:] Preheat oven to 450 F. Place pizza crust on baking sheet.
  \item[2:] Spread pesto over pizza crust. Arrange tomatoes over pesto. Sprinkle with cheese and crushed red pepper.
  \item[3:] Bake for 10 to 12 minutes or until cheese is melted and crust is golden brown. Cut into wedges.
\end{itemize}

\textbf{Question:} Can we \texttt{preheat} (Step 1) and \texttt{bake} (Step 3) at the same time? \\
\textbf{Instruction:} Please answer with only “Yes” or “No” without any further explanation. Adhere to this rule strictly.
\end{tcolorbox}
\caption{Prompt example for the Concurrency Analysis in the Recipes domain.}
\label{fig:recipes-concurrency-analysis}
\end{figure}

\clearpage
\newpage

\subsection{Domain-wise Results and Statistical Significance}

Table~\ref{tab:llm-domain-accuracy} presents the detailed accuracy scores of each evaluated model across the eight data-flow analyses in the three domains of \textsc{FABLE}. This format allows for fine-grained comparison within and across domains, highlighting how each model handles different procedural reasoning challenges. 

\begin{table}[ht]
\centering
\caption{Accuracy (\%) of each LLM across the eight FABLE data-flow analyses, grouped by domain.}
\resizebox{\textwidth}{!}{
\begin{tabular}{llrrr}
\toprule
\textbf{Domain} & \textbf{Analysis} & \texttt{deepseek-r1:8B} & \texttt{granite-code:8B} & \texttt{llama3.1:8B} \\
\midrule

\multirow{8}{*}{\textbf{Plans}} 
& Reaching Definitions   & 92 & 55 & 67 \\
& Very Busy Expressions  & 97 & 84 & 79 \\
& Available Expressions  & 87 & 89 & 55 \\
& Live Variable Analysis & 97 & 52 & 56 \\
& Interval Analysis      & 31 & 16 & 8  \\
& Type-State Analysis    & 94 & 47 & 51 \\
& Taint Analysis         & 100 & 27 & 42 \\
& Concurrency Analysis   & 63 & 62 & 52 \\
\midrule

\multirow{8}{*}{\textbf{Travel Routes}} 
& Reaching Definitions   & 99 & 48 & 62 \\
& Very Busy Expressions  & 100 & 35 & 48 \\
& Available Expressions  & 64 & 85 & 44 \\
& Live Variable Analysis & 84 & 48 & 46 \\
& Interval Analysis      & 39 & 0  & 2  \\
& Type-State Analysis    & 100 & 69 & 83 \\
& Taint Analysis         & 65 & 52 & 45 \\
& Concurrency Analysis   & 100 & 59 & 72 \\
\midrule

\multirow{8}{*}{\textbf{Recipes}} 
& Reaching Definitions   & 81 & 63 & 79 \\
& Very Busy Expressions  & 80 & 23 & 23 \\
& Available Expressions  & 75 & 77 & 32 \\
& Live Variable Analysis & 93 & 60 & 47 \\
& Interval Analysis      & 96 & 67 & 80 \\
& Type-State Analysis    & 86 & 46 & 85 \\
& Taint Analysis         & 80 & 35 & 73 \\
& Concurrency Analysis   & 80 & 34 & 49 \\

\bottomrule
\end{tabular}
}
\label{tab:llm-domain-accuracy}
\end{table}

To determine whether performance differences across domains are statistically meaningful, we conduct pairwise statistical significance tests for each model, comparing accuracy distributions across the three domains in \textsc{FABLE}. Specifically, we apply the two-tailed Welch’s t-test on the per-analysis accuracy values, which accounts for unequal variance and small sample sizes.

Let the domain-wise mean accuracy for a model \( M \) be denoted by \( \mu_d \), and standard deviation by \( \sigma_d \), where \( d \in \{\text{Plans}, \text{Travel}, \text{Recipes}\} \). Each accuracy vector has length \( n = 8 \), one value per data-flow analysis.

We compute the t-statistic for each pair of domains \( (d_1, d_2) \) using:
\[
t = \frac{\mu_{d_1} - \mu_{d_2}}{\sqrt{\frac{\sigma_{d_1}^2}{n} + \frac{\sigma_{d_2}^2}{n}}},
\]
with degrees of freedom \( \nu \) estimated via the Welch–Satterthwaite equation.

\begin{table}[ht]
\centering
\caption{Mean accuracy (\%) and standard deviation across the 8 data-flow analyses for each model in each domain.}
\label{tab:domain_stats}
\begin{tabular}{lccc}
\toprule
\textbf{Model} & \textbf{Plans} & \textbf{Travel Routes} & \textbf{Recipes} \\
\midrule
\texttt{deepseek-r1:8b}   & 82.6 ± 23.4 & 81.4 ± 25.6 & 83.9 ± 7.3 \\
\texttt{granite-code:8b}  & 54.0 ± 22.6 & 50.8 ± 18.1 & 43.6 ± 17.6 \\
\texttt{llama3.1:8b}      & 55.0 ± 18.2 & 50.3 ± 25.0 & 59.1 ± 22.4 \\
\bottomrule
\end{tabular}
\end{table}

\begin{table}[ht]
\centering
\caption{Welch’s t-test results (t-statistic and two-tailed p-value) comparing domain performance per model.}
\label{tab:t_test}
\begin{tabular}{lccc}
\toprule
\textbf{Model} & \textbf{Plans vs. Travel} & \textbf{Plans vs. Recipes} & \textbf{Travel vs. Recipes} \\
\midrule
\texttt{deepseek-r1:8b}   & \( t = 0.13,\, p = 0.897 \) & \( t = -0.14,\, p = 0.890 \) & \( t = -0.26,\, p = 0.799 \) \\
\texttt{granite-code:8b}  & \( t = 0.51,\, p = 0.623 \) & \( t = 1.08,\, p = 0.299 \) & \( t = 1.67,\, p = 0.123 \) \\
\texttt{llama3.1:8b}      & \( t = 0.33,\, p = 0.746 \) & \( t = -0.43,\, p = 0.678 \) & \( t = -0.67,\, p = 0.514 \) \\
\bottomrule
\end{tabular}
\end{table}

For all models, the p-values across domain comparisons are well above the standard threshold of \( \alpha = 0.05 \), indicating that no statistically significant differences exist in mean accuracy across domains. This result holds despite moderate differences in raw averages (e.g., \texttt{granite-code:8b} shows weaker performance on Recipes). However, the absence of statistical significance should not be interpreted as robustness. Rather, it reflects that performance variation across domains is comparable in magnitude to intra-domain variation across analyses. In particular, the high standard deviations (e.g., \( \sigma = 25.6 \) for \texttt{deepseek-r1:8b} in Travel) suggest considerable inconsistency in model behavior across different types of reasoning tasks.

These statistical results support the claim that LLMs tested on \textsc{FABLE} do not exhibit domain-specific reasoning strength or weakness in a statistically robust manner. More importantly, the consistently large variance across analyses highlights the diagnostic value of \textsc{FABLE} in surfacing fine-grained reasoning failures. Rather than punishing models for cross-domain shifts, the benchmark focuses attention on compositional, temporal, and causal generalization—challenges that remain open across all domains.

\subsection{Inference Time Analysis}

\begin{figure}[t]
    \centering
    \includegraphics[width=\linewidth]{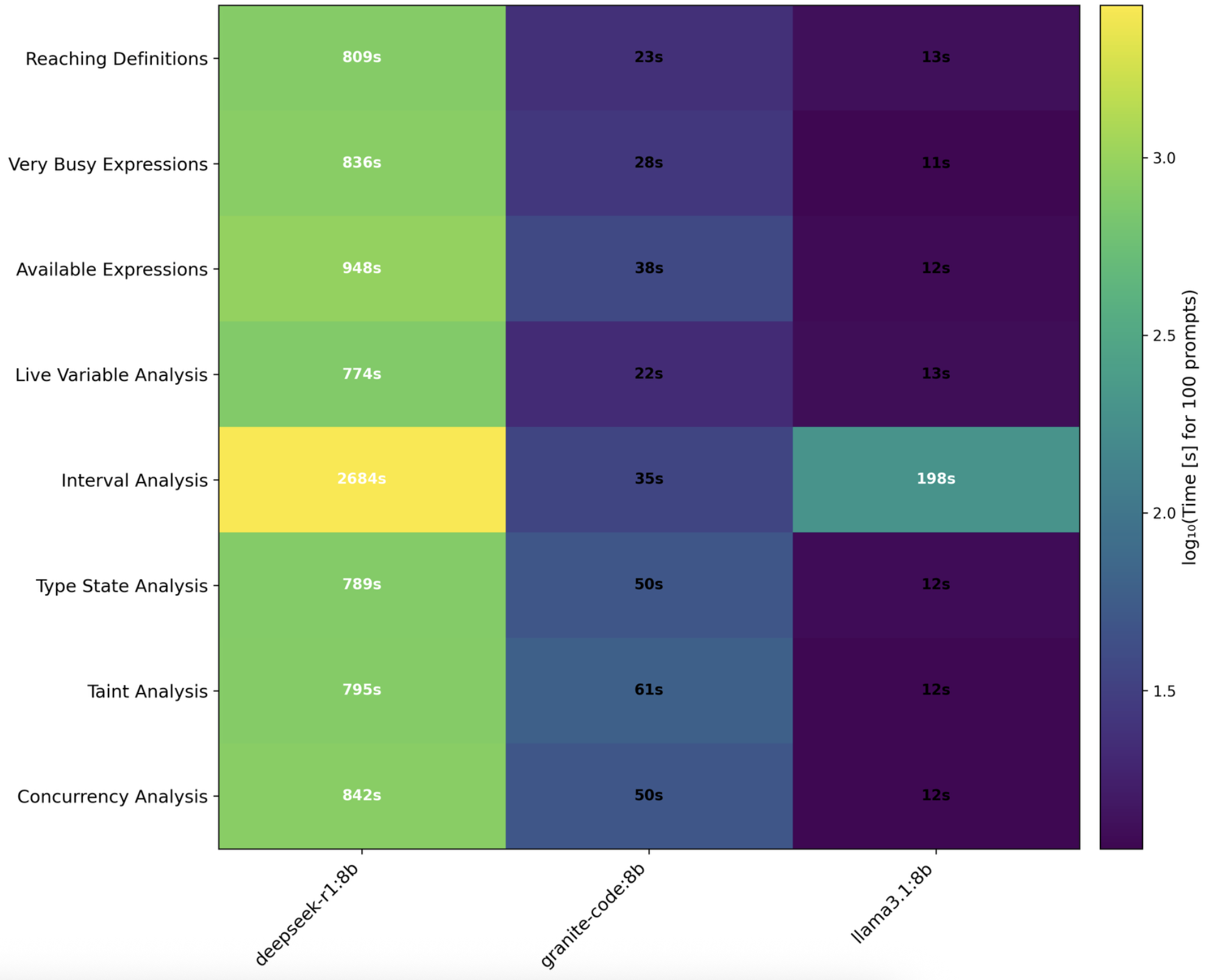}
    \caption{Inference latency heatmap for \textbf{Plans} domain. Color intensity indicates log-scale inference time (in seconds per 100 prompts) for each model–analysis pair.}
    \label{fig:inference-latency-plans}
\end{figure}

\begin{figure}[t]
    \centering
    \includegraphics[width=\linewidth]{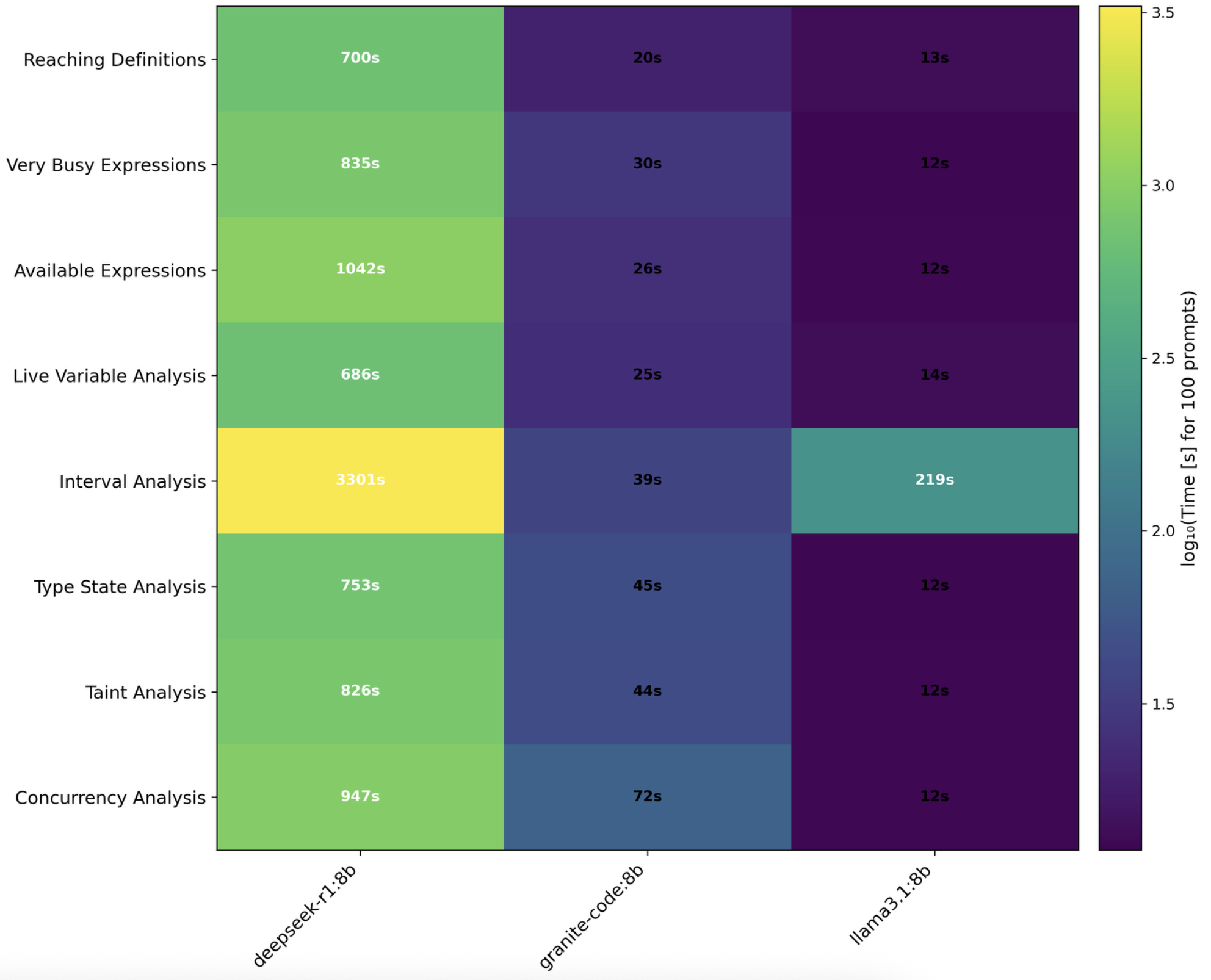}
    \caption{Inference latency heatmap for \textbf{Travel Routes} domain. Deep models like \texttt{deepseek-r1:8b} incur significantly higher latency, especially on Interval Analysis.}
    \label{fig:inference-latency-travel}
\end{figure}

\begin{figure}[t]
    \centering
    \includegraphics[width=\linewidth]{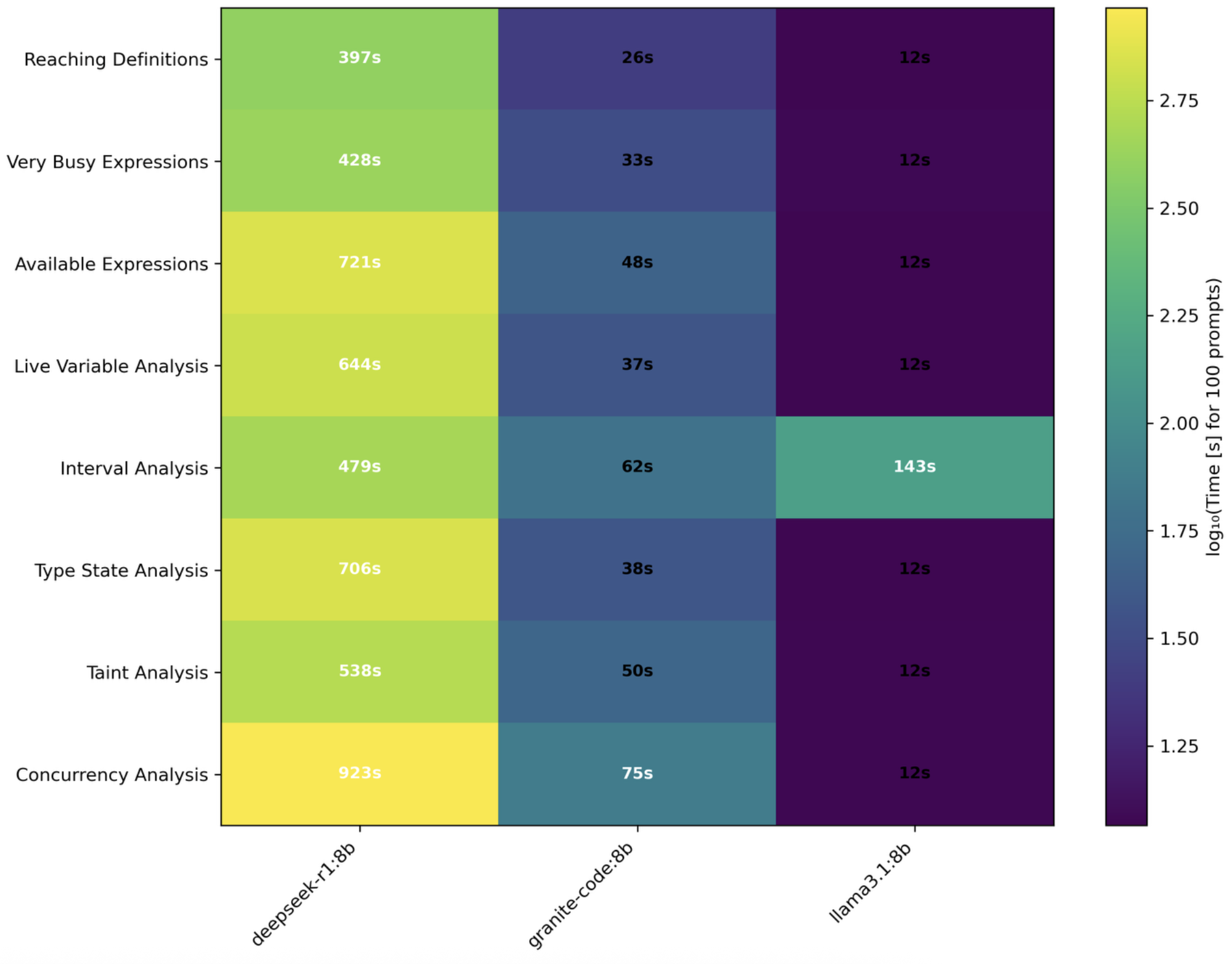}
    \caption{Inference latency heatmap for \textbf{Recipes} domain. All models show increased latency for Interval Analysis due to longer and more complex completions.}
    \label{fig:inference-latency-recipes}
\end{figure}

To complement accuracy-based evaluation, we analyze inference latency to assess computational efficiency across model classes. Figure~\ref{fig:inference-latency-plans}, \ref{fig:inference-latency-travel}, and \ref{fig:inference-latency-recipes} presents heatmaps of average generation time per 100 prompts for each data-flow analysis and domain, using a logarithmic color scale to accommodate large variance.

Across all domains, \texttt{deepseek-r1:8b} incurs significantly higher latency—often exceeding 700 seconds and peaking at over 3300 seconds for \textit{Interval Analysis} in Travel Routes—highlighting the computational overhead of its reasoning-focused architecture. In contrast, \texttt{granite-code:8b} and \texttt{llama3.1:8b} remain substantially faster, with most analyses completing within 20–75 seconds per 100 prompts. Notably, \textit{Interval Analysis} is consistently the most time-consuming across all models, reflecting the longer output spans and complex step-level parsing required for numeric-range and temporal inference.

These findings emphasize a critical tradeoff: higher-performing models like \texttt{deepseek-r1:8b} offer better accuracy but at an order-of-magnitude increase in compute time. Such latency bottlenecks raise important considerations for real-time deployment and call for future research into efficiency-aware model design and task-specific distillation strategies.

\end{document}